\documentclass[journal,comsoc]{IEEEtran}
\usepackage{url}
\usepackage{amsmath,graphicx,cite}
\usepackage{amsfonts}
\usepackage{authblk}
\usepackage{algpseudocode}
\usepackage[ruled,linesnumbered]{algorithm2e}
\usepackage{verbatim}
\usepackage{longtable}
\usepackage{rotating}
\usepackage{float}
\usepackage{soul}
\usepackage{multirow}
\usepackage{epstopdf}

\usepackage{tabularx}
\usepackage{bigints}
\usepackage{tikz}
\usepackage{multirow} 
\usepackage{amsmath,amstext,amsfonts,amssymb}
\usepackage{amsbsy}
\usepackage{amsthm}
\usepackage{diagbox}
\usepackage{mathrsfs}
\usepackage{bbm}
\usepackage{dsfont}
\usepackage{mathtools}
\usepackage{color}
\usepackage{caption}
\usepackage{subcaption}
\usepackage{morefloats}
\usepackage{float}
\usepackage[nolist]{acronym}
\usepackage{acronym}

\acrodef{itu}[ITU]{International Telecommunication Union}
\acrodef{6dof}[6DoF]{6 Degrees of Freedom}
\acrodef{sn}[SN]{Satellite Network}
\acrodef{an}[AN]{Aerial Network}
\acrodef{gn}[GN]{Ground Network}
\acrodef{rf}[RF]{Radio Frequency}
\acrodef{ris}[RIS]{Reconfigurable Intelligent Surface}
\acrodef{vlc}[VLC]{Visible Light Communications}
\acrodef{re}[RE]{Reflective Element}
\acrodef{geo}[GEO]{Geostationary Orbit}
\acrodef{meo}[MEO]{Medium Earth Orbit}
\acrodef{leo}[LEO]{Low Earth Orbit}
\acrodef{hap}[HAP]{High Altitude Platform}
\acrodef{lap}[LAP]{Low Altitude Platform}
\acrodef{vn}[VN]{Virtual Network}
\acrodef{uav}[UAV]{Unmanned Aerial Vehicle}
\acrodef{3d}[3D]{Three-Dimensional}
\acrodef{fso}[FSO]{Free-Space Optical}
\acrodef{owc}[OWC]{Optical Wireless Communications}
\acrodef{bert}[BERT]{Bidirectional Encoder Representations from Transformers}
\acrodef{bs}[BS]{Base Station}
\acrodef{ber}[BER]{Bit Error Rate}
\acrodef{thz}[THz]{Terahertz}
\acrodef{mmwave}[mmWave]{Millimeter Wave}
\acrodef{mimo}[MIMO]{Multiple-Input Multiple-Output}
\acrodef{gpt}[GPT]{Generative Pre-Trained Transformer}
\acrodef{dnn}[DNN]{Deep Neural Network}
\acrodef{cnn}[CNN]{Convolutional Neural Network}
\acrodef{gnn}[GNN]{Graph Neural Network}
\acrodef{vpcc}[V-PCC]{Video-based Point Cloud Compression}

\acrodef{los}[LoS]{Line-of-Sight}
\acrodef{nlos}[NLoS]{Non-Line-of-Sight}
\acrodef{ir}[IR]{Infrared}
\acrodef{uv}[UV]{Ultraviolet}
\acrodef{ioe}[IoE]{Internet-of-Everything}
\acrodef{iot}[IoT]{Internet-of-Things}
\acrodef{occ}[OCC]{Optical Camera Communication}
\acrodef{lifi}[LiFi]{Light Fidelity}
\acrodef{led}[LED]{Light-Emitting Diode}
\acrodef{ml}[ML]{Machine Learning}
\acrodef{qos}[QoS]{Quality-of-Service}
\acrodef{sagin}[SAGIN]{Space-Air-Ground Integrated Network}
\acrodef{csi}[CSI]{Channel State Information}
\acrodef{em}[EM]{Electromagnetic}
\acrodef{gan}[GAN]{Generative Adversarial Network}
\acrodef{snr}[SNR]{Signal-to-Noise Ratio}
\acrodef{sinr}[SINR]{Signal-to-Interference-plus-Noise Ratio}
\acrodef{e2e}[E2E]{End-to-End}
\acrodef{5g}[5G]{Fifth Generation}
\acrodef{mec}[MEC]{Mobile Edge Computing}
\acrodef{mems}[MEMS]{Microelectromechanical Systems}
\acrodef{6g}[6G]{Sixth Generation}
\acrodef{b5g}[B5G]{Beyond 5G}
\acrodef{tbps}[Tbps]{Tera-Bit-Per-Second}
\acrodef{llm}[LLM]{Large Language Model}
\acrodef{mllm}[MLLM]{Multimodal Large Language Model}
\acrodef{ios}[IoS]{Internet of Senses}
\acrodef{nlp}[NLP]{Natural Language Processing}
\acrodef{vlm}[VLM]{Vision Language Model}
\acrodef{vr}[VR]{Virtual Reality}
\acrodef{ar}[AR]{Augmented Reality}
\acrodef{mr}[MR]{Mixed Reality}
\acrodef{xr}[XR]{Extended Reality}
\acrodef{dt}[DT]{Digital Twin}
\acrodef{fpv}[FPV]{First-Person View}
\acrodef{bci}[BCI]{Brain-Computer-Interface}
\acrodef{fps}[FPS]{Frames Per Second}
\acrodef{hmd}[HMD]{Head Mounted Device}
\acrodef{psnr}[PSNR]{Peak Signal-to-Noise Ratio}
\acrodef{ai}[AI]{Artificial Intelligence}
\acrodef{ue}[UE]{User Equipment}
\acrodef{http}[HTTP]{Hypertext Transfer Protocol}
\acrodef{html}[HTML]{Hypertext Markup Language}
\acrodef{api}[API]{Application Programming Interface}
\acrodef{blip}[BLIP]{Bootstrapping Language-Image Pre-training}
\acrodef{qoe}[QoE]{Quality of Experience}
\acrodef{png}[PNG]{Portable Network Graphics}
\acrodef{genai}[GenAI]{Generative AI}
\acrodef{moe}[MoE]{Mixture of Expects}
\acrodef{fov}[FoV]{Field of View}
\acrodef{mqtt}[MQTT]{Message Queuing Telemetry Transport}
\acrodef{rtsp}[RTSP]{Real-Time Streaming Protocol}
\acrodef{avc}[AVC]{Advanced Video Coding}
\acrodef{webrtc}[WebRTC]{Web Real-Time Communication}
\acrodef{gltf}[GLTF]{Graphics Language Transmission Format}
\acrodef{glb}[GLB]{Binary GLTF}
\acrodef{kpi}[KPI]{Key Performance Indicator}
\acrodef{icl}[ICL]{In-Context Learning}
\acrodef{vit}[ViT]{Vision Transformer}
\acrodef{mulsemedia}[Mulsemedia]{Multi-Sensorial Media}
\acrodef{imu}[IMU]{Inertial Measurement Unit}
\acrodef{vod}[VoD]{Video on Demand}
\acrodef{dash}[DASH]{Dynamic Adaptive Streaming over HTTP}
\acrodef{hevc}[HEVC]{High Efficiency Video Coding}
\acrodef{erp}[ERP]{Equirectangular Projected}
\acrodef{rtmp}[RTMP]{Real-Time Messaging Protocol}
\acrodef{hls}[HLS]{HTTP Live Streaming}
\acrodef{json}[JSON]{JavaScript Object Notation}
\acrodef{lstm}[LSTM]{Long Short-Term Memory}
\acrodef{rnn}[RNN]{Recurrent Neural Network}

\usepackage{ulem} 
\usepackage{titlesec, booktabs }
\usepackage{hyperref, adjustbox}
\usetikzlibrary{arrows,shapes,positioning,shadows,trees}
\usetikzlibrary{decorations.pathreplacing}
\usepackage{pgfplots}
\usepackage{array}

 \usepackage{nicematrix}

\usepackage{graphicx}

\usepackage{comment}

\begin{document}
\renewcommand{\textcolor}[2]{#2}
\renewcommand{\color}[1]{}
\title{Generative AI for Immersive Communication: The Next Frontier in Internet-of-Senses Through 6G }  


\author{Nassim Sehad,~\IEEEmembership{Student~Member, IEEE}, Lina~Bariah,~\IEEEmembership{Senior~Member, IEEE}, Wassim Hamidouche,~\IEEEmembership{Senior~Member, IEEE}, Hamed Hellaoui, Riku Jäntti, ~\IEEEmembership{Senior~Member, IEEE}, and M{\'e}rouane Debbah,~\IEEEmembership{Fellow, IEEE}
} 






\vspace{-0.8cm}

\maketitle


\begin{abstract}
Over the past two decades, the \ac{iot} has \textcolor{blue}{become} a transformative concept, and as we approach 2030, a new paradigm known as the  \ac{ios} is emerging. Unlike conventional \ac{vr}, \ac{ios} seeks to provide multi-sensory experiences, acknowledging that in our physical reality, our perception extends far beyond just sight and sound; it encompasses a range of senses. This article explores \textcolor{blue}{the} existing technologies driving immersive multi-sensory media, delving into their capabilities and potential applications. This exploration includes a comparative analysis between conventional immersive media streaming and a proposed use case that leverages semantic communication empowered by generative \ac{ai}. The focal point of this analysis is the substantial reduction in bandwidth consumption by 99.93\% in the proposed scheme. Through this comparison, we aim to underscore the practical applications of generative \ac{ai} for immersive media. \textcolor{blue}{Concurrently} addressing \textcolor{blue}{major challenges in this field, such as temporal synchronization of multiple media, ensuring high throughput, minimizing the \ac{e2e} latency, and robustness to low bandwidth while} outlining future trajectories.
\end{abstract} 
\acresetall

\section{Introduction} 
The advent of the 5th generation (5G) mobile networks and recent advancements in computing technologies have redefined the concept of \textcolor{blue}{I}nternet from basic connectivity to a more advanced digital experience, transitioning from merely faster communication into an immersive interaction with the digital realm. This concept has been recently introduced under the umbrella of Metaverse and \textcolor{blue}{\acp{dt}.} \textcolor{blue}{It} has opened up a wide range of applications including \ac{vr}, \ac{ar}, holoportation, and teleoperation, among others. Within this realm, four main underpinnings have been remarked as paradigms for linking the cyber and physical worlds, namely, connected intelligent machines, a digitized programmable world, connected sustainable world, and the \ac{ios}~\textcolor{blue}{\footnote{\href{https://www.ericsson.com/en/reports-and-papers/consumerlab/reports/10-hot-consumer-trends-2030}{https://www.ericsson.com/en/reports-and-papers/consumerlab/reports/10-hot-consumer-trends-2030}}}. The \ac{ios} concept is set to revolutionize the digital interactions by creating a fully immersive environment that transcends traditional boundaries. By integrating sensory experiences such as sight, sound, touch, smell, and taste into the digital realm, this technology promises a more engaging cyber world, where virtual experiences are as rich and multi-dimensional as the physical world.

Human\textcolor{blue}{s} being experienc\textcolor{blue}{ing} the world through different senses, by perceiving sensory signals that are integrated or segregated in the brain. If these senses, especially haptic feedback, are accurately represented to be coherent with the real world, they can positively influence actions and behaviors, such as reaction time and detection~\cite{9143472}. Within this context, the \ac{ios} technology will allow individuals to experience a wide range of sensations remotely, revolutionizing various verticals, including industry, healthcare, networking, education, and tourism, to name a few. In order to reap the full potential of the \ac{ios} technology, numerous challenges need to be tackled to achieve a fully immersive multisensory experience. These challenges are pertinent to \textcolor{blue}{the} temporal synchronization of multiple media, addressing motion sickness, ensuring high throughput, and minimizing the \ac{e2e} latency. The collection of data from various sensor modalities, such as visual, audio, and haptic, plays a vital role in crafting a multisensory experience, in which this data can be synchronized at either the source or the destination (i.e., end devices or edge servers). The failure of virtual experiences to truly replicate our senses introduces confusion in human brains, leading to symptoms like nausea, dizziness, and migraines. To mitigate these drawbacks, it is crucial to enhance the realism of virtual sensations and reduce latency in  \ac{vr}/\ac{ar} devices, thereby minimizing latency between different modalities and avoiding its mismatch~\cite{pyun2022materials}. Furthermore, for accurate control purposes over a distance of up to one mile and to prevent the occurrence of motion sickness, it is crucial to transmit the sensory information at extremely low \ac{e2e} latency, ideally within 1-10 millisecond~\cite{fettweis2014tactile}. 

With respect to the \acp{kpi} for reliable communication of immersive media in \ac{ios}, it was demonstrated that future 6G networks should realize an E2E latency performance within the range of 1 ms for high-quality video streaming and haptic signals, with data rate requirements \textcolor{blue}{ranging from} tens \textcolor{blue}{of} Mbps to 1 Tbps and reliability performance of $10^{-7}$ \cite{akyildiz2023mulsemedia}. \textcolor{blue}{In addition}, while taste and smell signal requirements are less stringent than videos and haptics, it is essential to realize a perfect synchronization among signals from different senses to achieve the full potential of the \ac{ios}. Among various technologies, semantic communication emerges as a promising candidate \textcolor{blue}{for achiev\textcolor{blue}{ing} ultra-low latency communication through communicating the meanings/semantics of messages instead of communicating the physical signal, yielding faster and bandwidth-efficient transmission.}

\begin{figure*}[ht]
    \centering
\includegraphics[width=\textwidth]{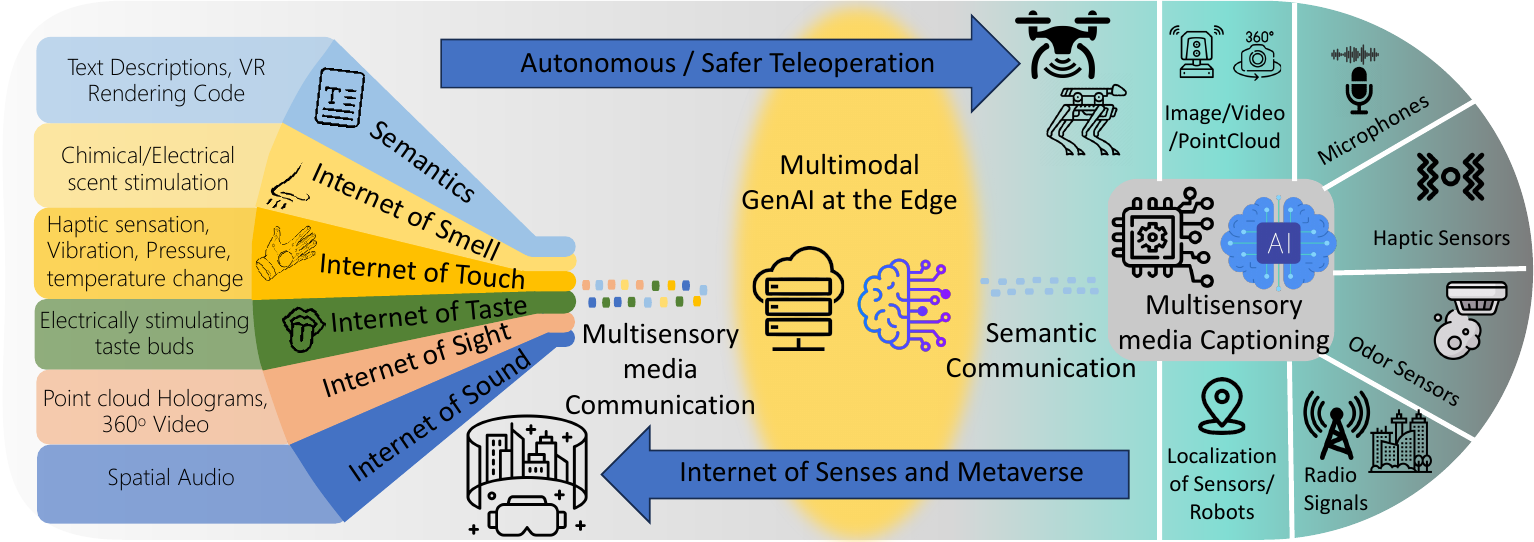}
    \caption{\textcolor{blue}{Key concepts of IoS}}
    \label{fig:1}

\end{figure*}

As advanced \ac{ai} systems, \acp{llm}, a subfield of \ac{ai}, was recently deemed as super-compressors that are capable of extracting the essential information to be communicated using a smaller message (a prompt) \cite{deletang2023language}. \Acp{llm} are \acp{dnn} with over a billion parameters, often reaching tens or even hundreds of billions, trained on extensive natural language datasets. This comprehensive parameterization unleashes a level of capability in generation, reasoning, and generalization that was previously unattainable in traditional \ac{dnn} models~\cite{NEURIPS2020_1457c0d6}. While the recovered messages by \ac{llm} will not be identical to the original one, they sufficiently represent their meanings and convey the intended messages. Accordingly, \acp{llm} are envisioned to evolve into the cognitive hub of the \ac{ios}, addressing intricate challenges like synchronization and compression \textcolor{blue}{by estimating  from partial modalities and enabling communication through semantic understanding.} Additionally, \acp{llm} are poised to enhance machine control intelligence, thereby improving reliability in teleoperations, through managing various data modalities pertinent to the user and environmental senses, as illustrated in Fig.~\ref{fig:1}. 

In recent developments, \acp{llm} have advanced to handle diverse modalities beyond text, encompassing audio, images, and video. The resulting \acp{mllm} can harness multiple data modalities to emulate human-like perception, integrating visual and auditory senses, and beyond~\cite{zhang2024mm}. \Acp{mllm} enable the interpretation and response to a broader spectrum of human communication, promoting more natural and intuitive interactions, including image-to-text understanding (e.g., \acs{blip}-2), video-to-text comprehension (e.g., LLaMA-VID), and audio-text understanding (e.g., QwenAudio). More recently, the development of \acp{mllm} has aimed at achieving any-to-any multi-modal comprehension and generation (e.g., VisualChatGPT).

In this paper, we aim to set the scene for the integration of \acp{llm} and the \ac{ios} technology, in which we develop a case study to demonstrate the benefits that can be obtained from exploiting the capabilities of \acp{llm} in enhancing the latency performance of immersive media communication. In particular, we conceptualize the 360$^{\circ}$ video streaming from a \ac{uav} as a semantic communication task. Initially, we employ object detection and image-to-text captioning to extract semantic information (text) from the input 360$^{\circ}$ frame. Subsequently, this generated textual information is transmitted to the edge server. 
\textcolor{blue}{In the edge server, an \ac{llm}  is utilized to produce WebXR code, facilitating the display of the corresponding image through \ac{3d} virtual objects on the \ac{hmd}, and estimate \ac{mulsemedia} sensors that actuate wearables to mimic the real environment's thermal and haptic sensations. Lastly, the generated \ac{mulsemedia} and code are sent to the receiver, allowing for the rendering of the \ac{3d} virtual content on the \ac{hmd} and direct actuation of haptic and thermal devices.} The contributions of this paper are summarized as follows: 
\begin{itemize}
\item Conceptualize the 360$^{\circ}$ video streaming via \ac{uav} as a semantic communication framework.
\item Harness the power of image-to-text captioning model and \ac{gpt} decoder-only \ac{llm} to generate A-frame code suitable for display on the user's \ac{hmd}.
\item Benchmark the proposed framework in terms of bandwidth consumption and communication latency across various components of the semantic communication framework.
\item Assess the quality of the generated \ac{3d} objects from our system compared to the captured 360$^{\circ}$ video images using reverse image-to-text, followed by text comparison through \ac{bert} model.


\end{itemize}

The remainder of this paper is organized as follows. Section~\ref{section2} introduces the \ac{ios} and discusses its necessity. In Section \ref{section3}, an overview of the development of  \acp{mllm} and their applications to  \ac{ios} is discussed. \textcolor{blue}{Section \ref{state} explores the state of the art in immersive media streaming.} Section \ref{section4} presents a case study with a proposed testbed, which is implemented and analyzed. \textcolor{blue}{Section \ref{results} presents the experimental results}. Finally, Section \ref{conclusion} highlights challenges and suggests directions for future research.

\section{Definitions and keys Concepts of  \ac{ios}}\label{section2}
In this section, we present the key concepts of the \ac{ios} concerning various interfaces and discuss the imperative nature of the \ac{ios}.

\subsection{Immersive All-Sense Communication}

To deliver a truly immersive experience, indistinguishable from reality, it is imperative to incorporate all human senses, including touch, taste, scent, as well as \acp{bci}, in addition to sight and sound. The human brain processes information from all senses to construct a comprehensive understanding of our environment. This necessity has given rise to the conceptualization of the \ac{ios}, a framework in which signals conveying information for all human senses are digitally streamed. This innovative concept aims to bridge the gap between physical and virtual reality, facilitating telepresence-style communication. Consequently, we categorize the various fundamental aspects of the \ac{ios} as the Internet of Touch, Internet of Taste, Internet of Smell, Internet of Sound, Internet of Sight, and \ac{bci}. Concurrently, Generative \ac{ai}, and more specifically, \acp{llm}, emerges as a pivotal concept within the \ac{ios} for semantic communication and synchronization. This is achieved by generating multiple media simultaneously, as illustrated in Fig.~\ref{fig:1}. \\\\
{ \bf Internet-of-Touch.} Haptic sensation refers to the sense of touch, known as tactile sensation, and it enhances immersive multimedia by allowing individuals to feel physical sensations, such as interactions with objects and movements (kinesthetic sensation). In \ac{vr} training or teleoperation, haptics replicate touch, which is crucial for tasks \textcolor{blue}{such as} surgery. Achieving optimal haptic experiences requires addressing minimal response times and low latency in synchronization with other sensed media, such as audio and video. Haptic interfaces employ various technologies to deliver tactile sensations, ranging from simple vibration feedback to more complex systems providing force feedback, pressure sensitivity, or even localized temperature changes. Devices \textcolor{blue}{including} haptic gloves, exoskeletons, or tactile feedback controllers enable users to touch, grasp, and interact with virtual objects in a natural and intuitive manner. \\\\
{\bf Internet-of-Taste.} Gustatory perception involves the intricate process of detecting and interpreting flavors. While traditional \ac{vr} primarily focuses on visual and auditory stimuli, incorporating taste into the virtual environment has the potential to enhance sensory engagement, leading to more realistic and immersive experiences. 
The technology underlying gustatory interfaces centers on the controlled stimulation of taste receptors. Various approaches are being explored, such as electrically stimulating taste buds~\cite{ranasinghe2012tongue} or delivering taste-related chemical compounds directly to the mouth.
However, it is crucial to note that replicating the sense of taste is the most complex, as it closely depends on other sensations. Presently, the technology is still in the laboratory demonstration stage. \\\\
{\bf Internet-of-Smell.} Digital scent technology, involved in recognizing or generating scents, employs electrochemical sensors and machine learning for scent recognition. Scent synthesis, on the other hand, utilizes chemical or electrical stimulation. Digital noses, electronic devices that detect odors, are increasingly prevalent in tasks such as quality control and environmental monitoring. In the food industry, digital noses ensure product quality by detecting off-flavors and maintaining taste and quality standards. In the perfume industry, digital noses evaluate aroma intensity and longevity, monitoring changes over time. Beyond industries, olfactory interfaces in everyday life enhance emotional and cognitive functions, productivity, and relaxation in virtual environments, as smell influences our daily emotions by 75\%~\cite{panagiotakopoulos2022digital}. This technology is particularly valuable in \ac{vr}, contributing to enhancing realism in training, enriching culinary experiences, evoking authentic atmospheres in tourism simulations, and aiding therapeutic applications. The technology behind smell interfaces involves the emission and dispersion of scents in a controlled manner. Different approaches have been explored, including the use of odor-releasing devices, cartridges, or even embedded scent generators within \ac{vr} headsets. These devices release scents or chemical compounds in response to specific cues or triggers, such as visual events or audio cues, to enhance the user's sensory experience. \\\\
{ \bf Internet-of-Sight.} 
\ac{xr} devices, encompassing \ac{vr}, \ac{ar}, and \ac{mr} headsets, glasses, or smart contact lenses, can offer a highly immersive experience for viewing video content along with haptic and other sensations. These devices have the capability to create a profound sense of presence and transport the viewer to a virtual environment, enabling them to feel as if they are physically present in the content. In recent years, the use of 360$^{\circ}$ video streaming has been on the rise, enabling viewers to experience immersive video content from multiple angles. This technology has gained popularity in various industries, including entertainment, sports, education, and robot teleoperation. \\\\
{\bf Internet-of-Audio.} Spatial audio pertains to the creation and reproduction of audio in a manner that simulates the perception of sound originating from various directions and distances. This process involves positioning sounds in a three-dimensional space to align with the visual environment. Spatial audio is a crucial element in crafting immersive experiences, as synchronized spatial audio reproduction complements visual information, thereby enhancing user immersion and \ac{qoe}~\cite{Choi_2019}. \\\\
{\bf The brain as a user interface.} \Acp{bci} enable direct communication and control by translating neural activity into machine-readable signals. In the context of the \ac{ios}, a brain is required to execute actions based on the perception of multiple senses. This can be either a human brain, utilizing a \ac{bci} for action, or a multimodal \ac{ai}.

\subsection{Why we need  \acs{ios}?}
The \ac{ios} holds significant potential in contributing to various technological advancements and enhancing user experiences in different domains. For example, in the entertainment domain, the heightened level of immersion can offer more realistic and engaging interactions, revolutionizing how users perceive and interact with digital content. Envisioning scenarios in movies, one not only witnesses but also smells the aftermath of an explosion, immersing oneself in the heat and vibrations of the scene. Furthermore, the \ac{ios} can contribute to advancements in healthcare by providing more accurate and real-time data for monitoring patients. For example, remote patient monitoring, telemedicine, and neuroimaging technologies can benefit from the \ac{ios} to improve diagnostics and treatment. At the business level, retail experiences can be enriched through multisensory interactions, and marketing strategies can achieve higher engagement by appealing to multiple senses. Also, with the \ac{ios}, the way humans interact with machines can become more intuitive and natural. Thought-controlled interfaces, allowing users to perform actions simply by thinking, have the potential to eliminate the need for traditional input devices and enhance the efficiency of human-machine interaction. Moreover, in hazardous situations and environments, workers can utilize telepresence technology enabled by the \ac{ios} to remotely control robots. This ensures safe operations in scenarios where the physical presence of humans could pose risks, such as handling dangerous materials or navigating challenging terrains.

\section{Foundation Models for \ac{ios}}
\label{section3}
In this section, we offer a concise overview of the evolution of foundation models towards \ac{mllm} and their potential applications in the era of the \ac{ios}, specifically focusing on image and video transmission.
\subsection {Advancement of Language Models} The progress in \ac{nlp} research has led to the development of models such as GPT-2, 
BART~\textcolor{blue}{\footnote{\href{https://huggingface.co/docs/transformers/en/model\_doc/bart}{https://huggingface.co/docs/transformers/en/model\_doc/bart}}}, and \ac{bert}~\textcolor{blue}{\footnote{\href{https://huggingface.co/docs/transformers/en/model\_doc/bert}{https://huggingface.co/docs/transformers/en/model\_doc/bert}}}
These models have sparked a new race to construct more efficient models with large-scale architectures, encompassing hundreds of billions of parameters. The most popular architecture is the decoder-only, including \acp{llm} like \ac{gpt}-3, 
Chinchilla and LaMDA. 
Following the release of open-source \acp{llm} like OPT 
and BLOOM, 
 more efficient open-source models have been recently introduced, such as Vicuna, 
Phi-1/2,  
LLaMa,  
FALCON,  
Mistral, 
and Mixtral\footnote{{
\href{https://huggingface.co/docs/transformers/model\_doc/mixtral}{https://huggingface.co/docs/transformers/model\_doc/mixtral}}}. This later follows the \ac{moe} architecture and training process initially proposed in MegaBlocks~\textcolor{blue}{\footnote{{
\href{https://huggingface.co/papers/2211.15841}{https://huggingface.co/papers/2211.15841}}}}.  Despite having fewer parameters, these models fine-tuned on high-quality datasets, have demonstrated compelling performance on various \ac{nlp} tasks, surpassing their larger counterparts. Furthermore, instruction tuning the foundation models on high-quality instruction datasets enables versatile capabilities like chat and code source generation, etc. The \acs{llm} have also shown unexpected capabilities of learning from the context (i.e., prompts), referred to as \ac{icl}.     

\subsection {Multimodal large language models}  Extending foundation models to multimodal capabilities has garnered significant attention in recent years. Several approaches of aligning visual input with the pre-trained \ac{llm} for vision-language tasks have been explored in the literature~\cite{xing2024survey}. Pioneering works such as VisualGPT
 and Frozen 
 utilized pre-trained \ac{llm} for tasks like image captioning and visual question answering. More advanced \acp{vlm} such as Flamingo, 
BLIP-2, 
and LLaVA 
follow a similar process by first extracting visual features from the input image using the CLIP \ac{vit} encoder. Then, they align the visual features with the pre-trained \ac{llm} using specific alignment techniques. For instance, LLaVA relies on a simple linear projection, Flamingo uses gated cross-attention, and BLIP-2 introduces the Q-former module. These models are trained on large image-text pair datasets, where only the projection weights are updated, and the encoder and the \ac{llm} remain frozen, mitigating training complexity and addressing catastrophic forgetting issues. 

In this era of \acp{mllm}, \ac{gpt}-4 has demonstrated remarkable performance in vision-language tasks encompassing comprehension and generation. Nevertheless, in addition to its intricate nature, the technical details of \ac{gpt}-4 remain undisclosed, and the source code is not publicly available, impeding direct modifications and enhancements.
To address these challenges, the MiniGPT-4 model was proposed. This model combines a vision encoder (\ac{vit}-G/14 and Q-Former) with the Vicuna \ac{llm}, utilizing only one projection layer to align visual features with the language model while keeping all other vision and language components frozen. The model is first trained on image-language datasets, then finetuned on high-quality image description pairs (3,500) to improve the naturalness of the generated language and its usability. The TinyGPT-V vision model, introduced by Yuan {\it et al.}~\cite{yuan2023tinygptv}, addresses computational complexity, necessitating only a 24GB GPU for training and an 8GB GPU for inference. The architecture of TinyGPT-V closely resembles that of MiniGPT-4, incorporating a novel linear projection layer designed to align visual features with the Phi-2 language model, which boasts only 2.7 billion parameters. The TinyGPT-V model undergoes a sophisticated training and fine-tuning process in four stages, where both the weights of the linear projection layers and the normalization layers of the language model are updated. As the process progresses, instruction datasets are incorporated in the third stage, and multi-task learning is employed during the fourth stage.

\begin{figure*}[ht]
    \centering
\includegraphics[width=\textwidth]{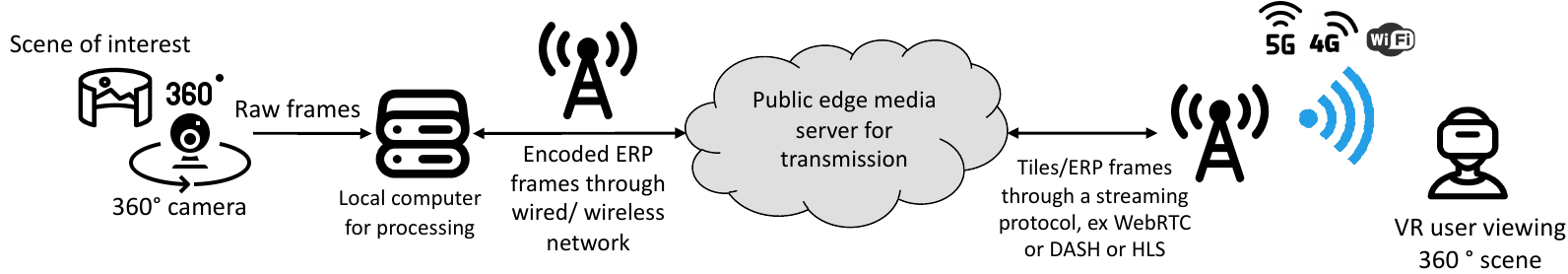}
    \caption{\textcolor{blue}{Architecture of a conventional video streaming system}}
    \label{fig:trad}
\end{figure*}
The second step in developing \acp{llm} is fine-tuning the model on instruction datasets to teach models to better understand
human intentions and generate accurate responses. The InstructBLIP~\textcolor{blue}{\footnote{{
\href{https://huggingface.co/docs/transformers/model\_doc/instructblip}{https://huggingface.co/docs/transformers/model\_doc/instructblip}}}} is built through instruct tuning of the pre-trained BLIP-2 model on 26 instruction datasets grouped into 11 tasks. During the instruction tuning process, the \ac{llm} and the image encoder are maintained frozen, while only the Q-former undergoes fine-tuning. Furthermore, the instructions are input to both the frozen \ac{llm} and the Q-Former. Notably, InstructBLIP exhibits exceptional performance across various vision-language tasks, showcasing remarkable generalization capabilities on unseen data. Moreover, when employed as the model initialization for individual downstream tasks, InstructBLIP models achieve state-of-the-art fine-tuning performance. InstructionGPT-4~\textcolor{blue}{\footnote{{
\href{https://huggingface.co/datasets/WaltonFuture/InstructionGPT-4}{https://huggingface.co/datasets/WaltonFuture/InstructionGPT-4}}}} is a vision model fine-tuned on a small dataset comprising only 200 examples, which represents approximately 6\% of the instruction-following data used in the alignment dataset for MiniGPT-4. The study highlights that fine-tuning the vision model on a high-quality instruction dataset enables the generation of superior output compared to MiniGPT-4.

\subsection {Potential Applications of \acp{mllm} in  Semantic Communication.} 
The lossy compression of images and videos has always involved a tradeoff between distortion ($D$), representing the reconstructed quality, and the coding rate ($R$). Distortion quantifies the errors introduced by compression, measured between the original sample ${\bf x}$ and its reconstructed version ${ \hat{\bf x}}$ as the p-norm distance $||{\bf x} - \hat{ \bf x}||^p_p$. The rate $R$ denotes the amount of data, in bits or bits/second, required to represent the sample after compression. Compression aims to minimize distortion under rate constraints, typically formulated as the minimization problem of the tradeoff between distortion and rate: $\min_{\bf \hat{x}} (D + \lambda R)$, where $\lambda$ is the Lagrangian parameter. 

In a real-time video transmission system, end-to-end latency plays a crucial role in determining system performance. Within this context, two distinct scenarios can be distinguished: offline video streaming and live video streaming. In the case of live video streaming, the end-to-end latency encompasses delays introduced by all streaming components, including acquisition, coding, packaging, transmission, depackaging, decoding, and display. Moreover, all these components need to operate at a frame frequency beyond the video frame rate. On the other hand, in the offline scenario, video encoding and packaging are performed offline. This exempts the process from real-time constraints and delays typically introduced by these two components. 


The recent advances in \ac{llm} and \ac{mllm} represent a transformative shift in video streaming. In this section, we explore three use cases integrating \ac{llm} and \ac{mllm} into the video streaming framework. The first use case involves the application of \ac{llm} for the lossless compression of images or videos, serving as an entropy encoder. Recent research, investigated by the work from DeepMind~\cite{deletang2023language}, underscores the potent versatility of \acp{llm} as general-purpose compressors, owing to their in-context learning capabilities. Experiments utilizing Chinchila 70B, solely trained in natural language, revealed impressive compression rations, achieving 43.4\% on ImageNet patches. Notably, this rate outperforms domain-specific image compressors such as \ac{png} (58.5\%). 

The second use case harnesses \ac{mllm} shared at both the transmitter and receiver for a lossy coding setting. The transmitter first generates an accurate description of the image or video content through the image captioning capability of the \ac{mllm}. Instead of transmitting the image or video, the text description (semantic information) is then sent to the receiver, requiring a significantly lower data rate. At the receiver, the generative capability of the \ac{mllm} is harnessed to reconstruct the image or video based on the received text description.

In the third use case, the \ac{mllm} is employed solely at the transmitter to leverage its code-generation capability, representing the image or video for transmission. Subsequently, the code, requiring a lower data rate, is shared with the receiver, enabling direct utilization to render the image or video through the code description. The intricacies of this latter use case are expounded upon and experimentally explored in the subsequent sections of this paper.



\section{State of the Art of conventional and semantic immersive media streaming Methods}
\label{state}

\begin{table*}[ht]
\caption{Comparison of Different Papers on Streaming Technologies}
\adjustbox{max width=\textwidth}{
\centering

\textcolor{blue}{
\begin{tabular}{c|l|c|c|c|c|c}
    \toprule
    {\textbf{  Category}} & {\textbf{ Paper}} & {\textbf{  Type}} & {\textbf{ Media Type}} & {\textbf{ Streaming Protocol}} & {\textbf{ Encoder / Decoder}} & {\textbf{ Design Objective}} \\
    \midrule
    \multirow{8}{*}{{ \rotatebox[origin=c]{90}{\hspace{-1.4cm}  Conventional Live Streaming}}}
    & { Lo et al. (2018) \cite{lo2018edge}} & { \acs{vod}} & { 360$^{\circ}$ video \acs{erp} frames} & { \acs{dash}} & { \acs{hevc}/H.265 codec} & { Bandwidth and latency} \\
      \cmidrule(r){2-7} 
    & { Taleb et al. (2022) \cite{taleb2022vr}} & { Live} & { 360$^{\circ}$ video \acs{erp} frames} & { \acs{webrtc}} & { \acs{avc}/H.264 codec} & { Ultra low latency} \\
      \cmidrule(r){2-7} 
    & { Chen et al. (2021) \cite{chen2021optimized}} & { Live} & { 360$^{\circ}$ video \acs{erp} tiles} & { \acs{dash}} & { \acs{avc}/H.264 codec} & { Bandwidth $>$ 50\% traditional streaming} \\
      \cmidrule(r){2-7} 
    & { Yi et al. (2020) \cite{yi2020analysis}} & { Live} & { 360$^{\circ}$ video \acs{erp} frames} & { \acs{rtmp} over \acs{http}-FLV} & { \acs{avc}/H.264 codec} & { Latency $<$ 4s for 1440p resolution} \\
     \cmidrule(r){2-7} 
    & { Park et al. (2023) \cite{park2023omnilive}} & { Live} & { Spherical to 360$^{\circ}$ video \acs{erp}} & { \acs{rtmp} over \acs{hls}} & { \acs{avc}/H.264 codec} & { Super resolution and bandwidth saving} \\
    \cmidrule(r){2-7} 
    & { Gao et al. (2024) \cite{gao2024low}} & { Live} & { 360$^{\circ}$ video \acs{erp} tiles} & { \acs{rtmp} over LL-\acs{dash}} & { \acs{avc}/H.264 codec} & { Scalability} \\
   \cmidrule(r){2-7} 
    & { De Fr{\'e} et al. (2024) \cite{de2024scalable}} & { Live} & { Head position + \acs{3d} video} & { \acs{webrtc}} & { Draco codec} & { 360ms latency for 1080p video at 3Mb/s} \\
   \cmidrule(r){2-7} 
    & { Us{\'o}n et al. (2024) \cite{uson2024untethered}} & { Live} & { Volumetric video} & { \acs{webrtc}} & { \acs{vpcc} codec} & { Optimal latency at 70Mb/s bandwidth} \\
   \midrule
    \multirow{5}{*}{{ \rotatebox{90}{\hspace{-1.4cm}  Semantic Streaming}}} 
    & { Xia et al. (2023) \cite{xia2023wiservr}} & { \acs{vod}} & { 360$^{\circ}$ video tiles} & { /} & { \acs{cnn} Encoder / Decoder} & { Reduce latency with reliable transmission} \\
   \cmidrule(r){2-7} 
    & { Ahn et al. (2024) \cite{ahn2024dynamic}} & { \acs{vod} } & { Video over text semantics} & { /} & { GPT-4 Encoder / DALLE-2 Decoder} & { Video content creation} \\
   \cmidrule(r){2-7} 
    & { Chen et al. (2024) \cite{chen2024cross}} & { \acs{vod}} & { Text, audio, image, haptics} & { /} & { \begin{tabular}[c]{@{}c@{}}\acs{gnn} Encoder/\acs{3d} generative \\ reconstruction network Decoder\end{tabular}} & { \acs{3d} object construction} \\
  \cmidrule(r){2-7} 
    & { Du et al. (2023) \cite{du2023yolo}} & { \acs{vod}} & { \acs{3d} objects through text} & { /} & { \acs{cnn} (YOLOv7) Encoder / Database Decoder} & { Optimize transmission power} \\
   \cmidrule(r){2-7} 
    & { Ours} & { Live} & {\begin{tabular}[c]{@{}c@{}} Text, video frames, temperature,\\  and haptics \end{tabular}} & { \begin{tabular}[c]{@{}c@{}}\acs{http} for semantics, \acs{mqtt} for \\ sensorial data and generated code\end{tabular}} & \begin{tabular}[c]{@{}c@{}} \acs{cnn} + \acs{rnn} + GPT-3.5 Encoder \\ GPT-4 Decoder  \end{tabular} & {  Optimize bandwidth consumption} \\
    \bottomrule 
\end{tabular}
}
}
\label{tab:stateofart}
\end{table*}


 \textcolor{blue}{The latest implementations and research on live immersive media streaming typically adhere to the conventional pipeline illustrated in Fig.~\ref{fig:trad}. This pipeline involves capturing a scene using either a 360$^{\circ}$ camera or multiple cameras, followed by stitching the frames and encoding. The encoding can occur either at the camera itself or on a separate processing unit. Subsequently, the frames are projected into an \ac{erp} format or cube map and encoded using traditional video standards such as \acs{avc}/H.264 or \acs{hevc}/H.265. Due to the resource limitations of 360$^{\circ}$ cameras, the encoded stream is usually transmitted to a remote media server using \ac{rtmp} or \ac{rtsp}. The media server may then re-encode the video before transmitting it to the end-user via \ac{dash}, \ac{webrtc}, or another media streaming protocol. Previous studies have shown \ac{webrtc} to be particularly effective due to its ultra-low latency and adaptive bitrate capabilities~\cite{sharma2023uav}. For \ac{vod} services, the primary distinction lies in the storage of the video on a cloud server instead of real-time transmission. For point cloud video, alternative codecs such as Google Draco or \ac{vpcc} are employed.}.

\textcolor{blue}{Recent research has explored the use of \ac{ai} as a compressor for semantic communication, transmitting only essential knowledge and information for scene reconstruction at the receiver. This approach holds promise for reducing redundant data and conserving bandwidth, proving particularly advantageous in high-mobility, frequent-handover scenarios like \ac{uav} communication and control. 
Table~\ref{tab:stateofart} summarizes recent studies comparing traditional streaming pipelines with semantic communication approaches in terms of protocols, codecs, and design.  While some traditional techniques incorporate \ac{fov} prediction and tile encoding for bandwidth optimization, they still operate in the megabits per second range. This limitation can result in video feed loss in severely bandwidth-constrained environments, a challenge not yet addressed by existing methods. Furthermore, current semantic communication-based solutions often remain confined to simulations and are not tailored for real-time applications.
Our proposed architecture, to our knowledge, is the first \ac{genai}-based encoder/decoder for immersive multimedia streaming in a real-time, ultra-low latency application like \ac{uav} control.  We have chosen the solution in \cite{taleb2022vr} as a benchmark because it represents one of the optimal traditional pipelines based on \ac{webrtc}, achieving ultra-low \ac{e2e} latency $\leq$ 600ms) for immersive streaming in \ac{uav} control scenarios.}

\begin{figure*}[ht]
    \centering
\includegraphics[width=\textwidth, height=3.3in]{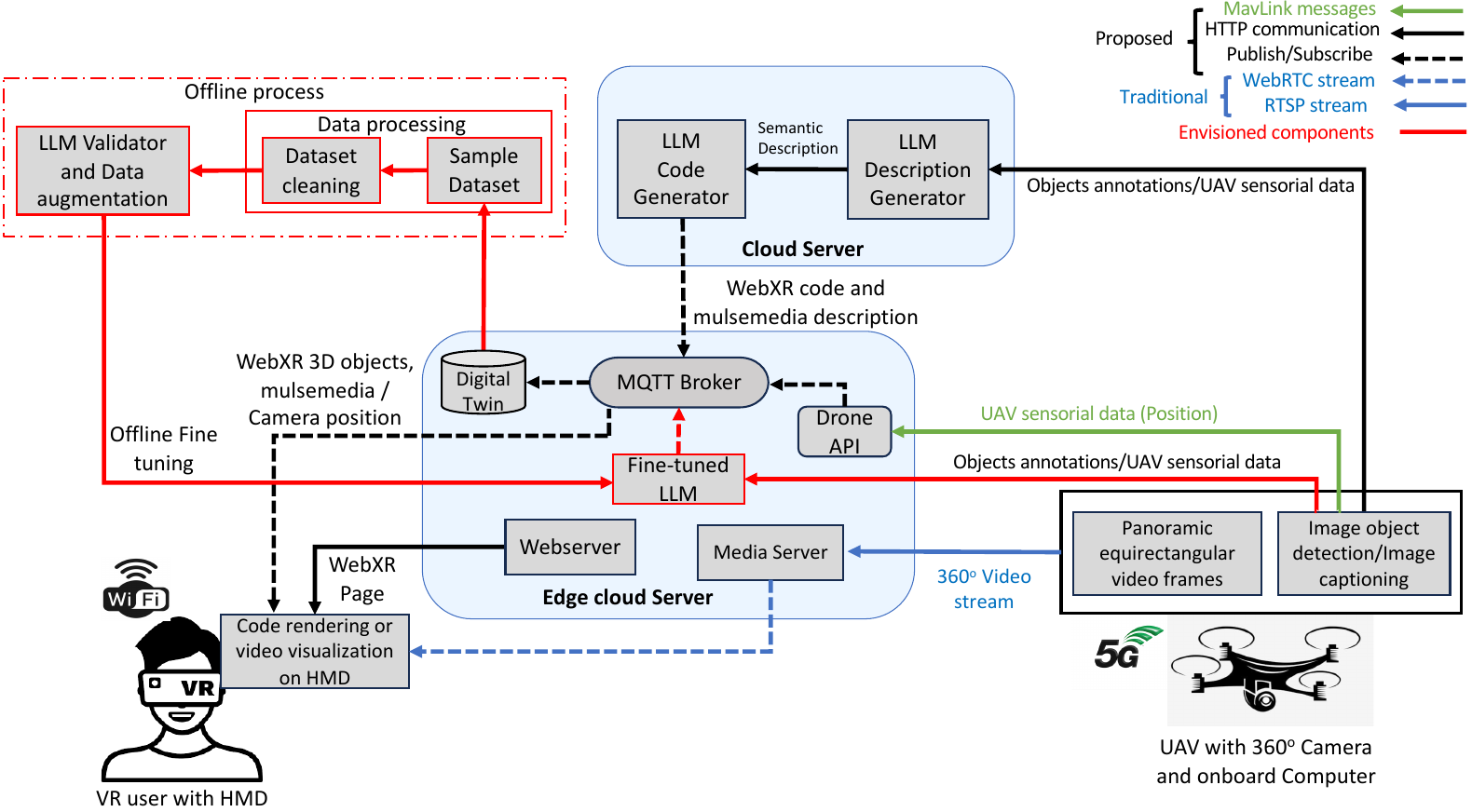}
    \caption{\textcolor{blue}{Proposed architecture for \ac{genai} enabled immersive communication}}
    \label{fig:2}
    \vspace{-0.3cm}
\end{figure*}

\section{Case Study} 
\label{section4}
\subsection{Use case description} 
{\color{blue}To comprehend the challenges at hand and explore potential solutions, let us immerse ourselves in the following scenario. 
John, a surveillance teleoperator, is tasked with remotely piloting a drone through a dense forest using a \ac{fpv} system over a  \ac{b5g} network. The task of navigating this complex environment through \ac{fpv} poses significant difficulties due to two primary factors:

\begin{itemize}
\item Limited Bandwidth: The forest environment inherently restricts bandwidth, leading to a degraded video stream in John's \ac{fpv} system. This degradation impairs his ability to effectively control the drone, potentially resulting in hazardous situations.
\item Limited Sensory Input: The drone's 360-degree camera, while providing visual and auditory feedback through the \ac{fpv} system, fails to fully capture the rich sensory context of the \ac{uav}'s surroundings. Achieving a truly immersive and comprehensive understanding of the drone's environment would require additional sensory inputs beyond the traditional visual and auditory sensors.
\end{itemize}

To address these challenges, we propose an architecture that leverages \ac{genai} for semantic communication. This approach aims to:

\begin{itemize}
    \item Reduce Bandwidth Consumption: \ac{genai}'s code-generation capabilities can be used to replicate the drone's video feed, minimizing bandwidth usage. This provides John with a secondary video stream, ensuring continuous operation even if the primary stream is interrupted.
    \item Enhance Sensory Immersion: \ac{genai} can generate additional sensory information beyond the traditional visual and audio streams, paving the way for an \ac{ios} experience. This will allow John to perceive the environment more comprehensively, improving his ability to control the drone safely and effectively.
\end{itemize}

By implementing this architecture, we can create a more immersive and reliable remote drone operation system, enabling teleoperators to navigate challenging environments with greater confidence and precision.
}

\textcolor{blue}{Furthermore, \ac{dt} based on \ac{3d} simulated environments have received a lot of interest from researchers, specifically for \ac{uav} teleoperation. \cite{meng2023dtuav} proposes a \ac{dt} framework for \ac{uav} monitoring and autonomy in which the \ac{uav} executes missions only after successful simulation of the \ac{uav} in the \ac{dt}. 
\cite{10437527} present a framework for \ac{uav} control through \ac{vr} comprising a \ac{dt} \ac{uav} equipped with virtual sensors that override user commands if obstacles in the \ac{dt} environment are detected nearby, thus providing reliable teleportation.}
\textcolor{blue}{
However, all of those \ac{dt}-based solutions rely on static \ac{3d} maps, which tend to evolve over time with the incorporation of temporary objects, thus making the static \ac{dt} unreliable. Therefore, we solve the latter issue by leveraging our proposed architecture. We enable \ac{uav}s to capture detailed environmental data, specifically of temporary elements in the environment that are difficult to find in any \ac{3d} database.  
Thus, we are able to inject the temporary objects generated by \ac{genai}, as proposed in our framework, into the \ac{dt}, and save bandwidth by streaming only the changing elements of the environment.} 


\subsection{Proposed Architecture for GenAI Enabled Immersive Communication}
The proposed architecture empowers a \ac{vr} user to visualize animated \ac{3d} digital objects crafted using WebXR code generated from \ac{llm}. The code for the \ac{3d} objects is generated based on feedback from the \ac{uav}'s mounted 360$^{\circ}$ camera, capturing omnidirectional frames of the environment. It is noteworthy that the \ac{3d} objects are rendered while the \ac{vr} user teleoperates the \ac{uav}. Simultaneously, the traditional method of transmitting 360$^{\circ}$ videos is employed as a baseline for comparison, considering factors such as delay, bandwidth consumption, and video quality achieved by our approach.

\textcolor{blue}{Beyond animated \ac{3d} objects, the \ac{llm} is able to estimate temperature along wind's speed and direction thereafter generate a \ac{mulsemedia} values map to activate the vibrators of a wearable haptic suit such as the Teslasuit~\textcolor{blue}{\footnote{\href{https://teslasuit.io/products/teslasuit-4/}{https://teslasuit.io/products/teslasuit-4/}}} and its thermal sensors based on the built-in sensors of the \ac{uav}'s flight controller, such as altitude, speed, and \ac{imu}. These estimations enable precise replications of environmental conditions, including tactile and thermal feedback, synchronized with the generation of the \ac{3d} objects. This creates a comprehensive \ac{ios} experience, allowing the user to feel the environmental conditions in real-time, enhancing the immersive quality of the virtual environment.}

The architecture is grounded in an edge-to-cloud continuum environment, as illustrated in Fig.~\ref{fig:2}. The individual components of the end-to-end video streaming architecture are elaborated upon in detail below. \\\\
{\bf \ac{vr} user.} The \ac{vr} user is an individual teleoperator, managing one or multiple \acp{uav} in a \ac{fpv} mode using a \ac{hmd} and its joysticks. The viewing interface is a WebXR-rendered webpage hosted on a web server operating on an edge cloud located in close proximity to the \ac{vr} user. This web server serves two pages: one displaying the 360$^{\circ}$ view and another presenting \ac{3d} objects from the environment.

Concurrently with the \ac{3d} view, the \ac{vr} headset and other wearables receive a \textcolor{blue}{\ac{mulsemedia}} description file containing values related to temperature and vibration. These values are derived from the environmental image and \ac{uav} movements, estimated using the \ac{llm}. The haptic feedback and potential heat dissipation wearable replicate the estimated environment concurrently with the view, minimizing synchronization latency as much as possible. Notably, since the generated code represents an animation, it is not necessary to update the view at a high frequency. However, the virtual camera in the spherical projection moves according to the drone's position, continuously received from the \ac{mqtt} broker by the user. \\\\
{ \bf Unmanned Aerial Vehicle.} The \ac{uav} functions as the real-world data capture device, employing its mounted 360$^{\circ}$ camera to provide live feedback in the form of a 360$^{\circ}$ video to the \ac{vr} user. Simultaneously, the \ac{uav}'s \textcolor{blue}{onboard computer} performs object detection \textcolor{blue}{on the camera's captured frames} using YOLOv7, trained on the MS COCO~\textcolor{blue}{\footnote{\href{https://cocodataset.org/}{https://cocodataset.org/}}} Dataset, and subsequent captioning using Inception-v3 \textcolor{blue}{and \ac{lstm}}~\cite{degadwala2021image} on one frame every 30 frames. The resulting object annotations \textcolor{blue}{resulting from the object detection, brief caption from the \ac{lstm},} and sensorial data \textcolor{blue}{from the \ac{uav}'s embedded sensors}, including position \textcolor{blue}{data}, are transmitted to the cloud \textcolor{blue}{in a \ac{json} format}, specifically to a \ac{llm} for additional contextualization and detailed description. Furthermore, \textcolor{blue}{in parallel} the \ac{uav} independently streams its position data, comprising altitude, latitude, and longitude, through Mavlink telemetry messages to a drone \ac{api}. This information is then relayed to the \ac{hmd} via the \ac{mqtt} broker. 
It is important to highlight that we are not hosting an \ac{llm} on the \ac{uav} due to its substantial size, computing demands, and energy consumption. Doing so would significantly reduce flight times.
\\\\
{ \bf Cloud server.} The cloud server primarily hosts \ac{http} \acp{api} connected to two \acp{llm}, specifically the first \ac{llm}, \textcolor{blue}{ GPT-3.5-Turbo, which has 175 billion parameters, and the second \ac{llm}, GPT-4, which has about 1.8 trillion estimated parameters.} 
Consequently, the first \ac{llm} is responsible for providing more context (enhanced image captioning) from an image caption and its object annotation, received from the \ac{uav}. 
Subsequently, the second \ac{llm} is prompted with the generated description from the first \ac{llm} as instruction and is tasked with generating \ac{html} code using the A-frame~\textcolor{blue}{\footnote{\href{https://aframe.io/}{https://aframe.io/}}} framework to produce immersive \ac{3d} WebXR content representing the image description in a \ac{3d} space. 

It is worth noting that we \textcolor{blue}{employ a multi-agent architecture with two distinct \acp{llm}, each assigned a specific task to enhance the accuracy of responses especially that \acp{llm} might not perform well when dealing with longer text sequences or tasks that require long term planning.} 
This strategy avoids directly feeding captions from the \ac{uav} to the second \ac{llm}. Instead, the first \ac{llm} fuses captions, annotations, and \ac{uav} sensor data, resulting in more detailed captions compared to standard ones 
Furthermore, by leveraging the prompt history stored in the \ac{llm} memory, we enhance the accuracy of the descriptions through the \ac{llm}'s in-context learning capability. \textcolor{blue}{Subsequently, we utilize a second \ac{llm} for code generation, ensuring that it does not impact the memory context of the first \ac{llm}. This multi-agent approach has been shown to improve response accuracy, with potential enhancements exceeding 6\% for GPT-3.5.}\\\\ 
\noindent {\bf Edge Cloud.} The edge cloud, located in close proximity to the drone, plays a crucial role in three fundamental computations: video streaming and transcoding, message transmission through a publish/subscribe broker, and web serving. In the traditional method of streaming 360$^{\circ}$ videos, a media server is utilized to receive an \ac{rtsp} video stream of equirectangular projected frames. Subsequently, these frames are transcoded using an \ac{avc}/H.264 encoder for re-streaming through \ac{webrtc}, as illustrated in~\cite{taleb2022vr}.

In contrast, in our proposed architecture centered on Generative AI-driven semantic communication, we employ WebXR code generated using the \ac{llm}, specifically \ac{gpt}-4, to represent virtual \ac{3d} objects and multimodal descriptions. This data is then transmitted to the user through a \ac{mqtt} broker and stored in the edge server to construct a \ac{dt} of the environment. An important consideration is that we refrain from hosting the \acp{llm} at the edge due to their large size and computing requirements.\\\\
{\bf Envisioned components.}
Notably, the optimal scenario aims to run all processes near the end user and \ac{uav}, reducing delays and eliminating the need for a separate cloud server. However, in our specific case, this has not been implemented due to limitations in the power of the edge server. These limitations are inherent in edge devices, rendering them insufficient for running an \ac{llm} with 70 billion parameters. Consequently, the proposed solution involves creating a fine-tuned version of the \ac{llm} that is suitable for hosting on the edge server. The procedure for developing this enhanced \ac{llm} is detailed in the workflow of our proposed architecture and further explained below.

In the existing workflow, the generated code is stored in the edge server within the \ac{dt} component. To create a fine-tuned model, supervised fine-tuning is required using both the prompt and the corresponding output of the \ac{llm}. This data must undergo thorough cleaning to eliminate redundancy and errors. Errors can be identified and corrected using another \ac{llm}, which then augments our dataset by generating similar data. Once we have this refined dataset of prompt pairs, it can be utilized to develop a quantized fine-tuned model that is capable of running directly on the edge server. This approach aims to further minimize communication latency.

\section{Experimental results}
\label{results}
In this section, we present the experimental test based on the architecture depicted in Fig.~\ref{fig:2}, along with measurements,  results analysis, and validation.

\subsection{Experimental setup} 

The experiment entailed a flight test conducted in proximity to Aalto University. Equirectangular videos, coupled with authentic footage captured by the 360$^{\circ}$ camera affixed to the \ac{uav}, were streamed to the \ac{vr} user situated at Aalto University. This streaming process was carried out via both the conventional approach and our novel method. Throughout the experiment, the \ac{uav} predominantly maneuvered at various altitudes while adhering to a maximum speed of 5m/s. \\

\noindent {\bf Video sequences.} For the experiments, we utilized 9 video sequences\footnote{\href{https://www.mettle.com/360vr-master-series-free-360-downloads-page/}{https://www.mettle.com/360vr-master-series-free-360-downloads-page/}}, boasting a 4K resolution, and streamed by an onboard computer, as detailed in Table~\ref{table:2}. We subjected these 9 videos to tests and measurements to assess bandwidth consumption and latency. Furthermore, we employed an additional video for validation purposes, evaluating description similarity results. Notably, the 10\textsuperscript{th} custom video, recorded by our team, underwent evaluation during a flight test conducted with the \ac{uav} at Aalto University. 
\begin{table}[ht]
\caption{Description of Videos and Their Durations}
\begin{tabular}{p{0.1\linewidth}p{0.5\linewidth}p{0.25\linewidth}}
\toprule
Video & Video Description & Duration \\
\midrule
1 & Thailand Stitched 360$^{\circ}$ footage & 25s \\
2 & Pebbly Beach & 2mins \\
3 & Bavarian Alps & 2.05mins \\
4 & Crystal Shower Falls & 2mins \\
5 & London on Tower Bridge & 30s \\
6 & London Park Ducks and Swans & 1.05mins \\
7 & View On Low Waterfall with Nice City & 10s \\
8 & Doi Suthep Temple & 25s \\
9 & Ayutthaya \ac{uav} Footage & 35s \\
10 & \ac{uav} video of Aalto University Finland & 2mins \\
\bottomrule
\label{table:2}
\end{tabular}
\vspace{-0.7cm}

\end{table}

\noindent {\bf Network and used hardware.} Following the global architecture, the testbed comprises a \ac{uav} equipped with a 5G modem for communication, an edge \ac{hmd} with a Wi-Fi connection (chosen due to the operator's indoor location), an edge cloud \textcolor{blue}{server} connected through fiber, and a cloud server connected via fiber. All components are located in Finland within distances less than 1 km from each other, except for the cloud server situated in the USA. Table~\ref{table:1} provides a detailed description of the hardware used. 

\begin{figure}[ht]
    \centering
    \includegraphics[width=0.48\textwidth]{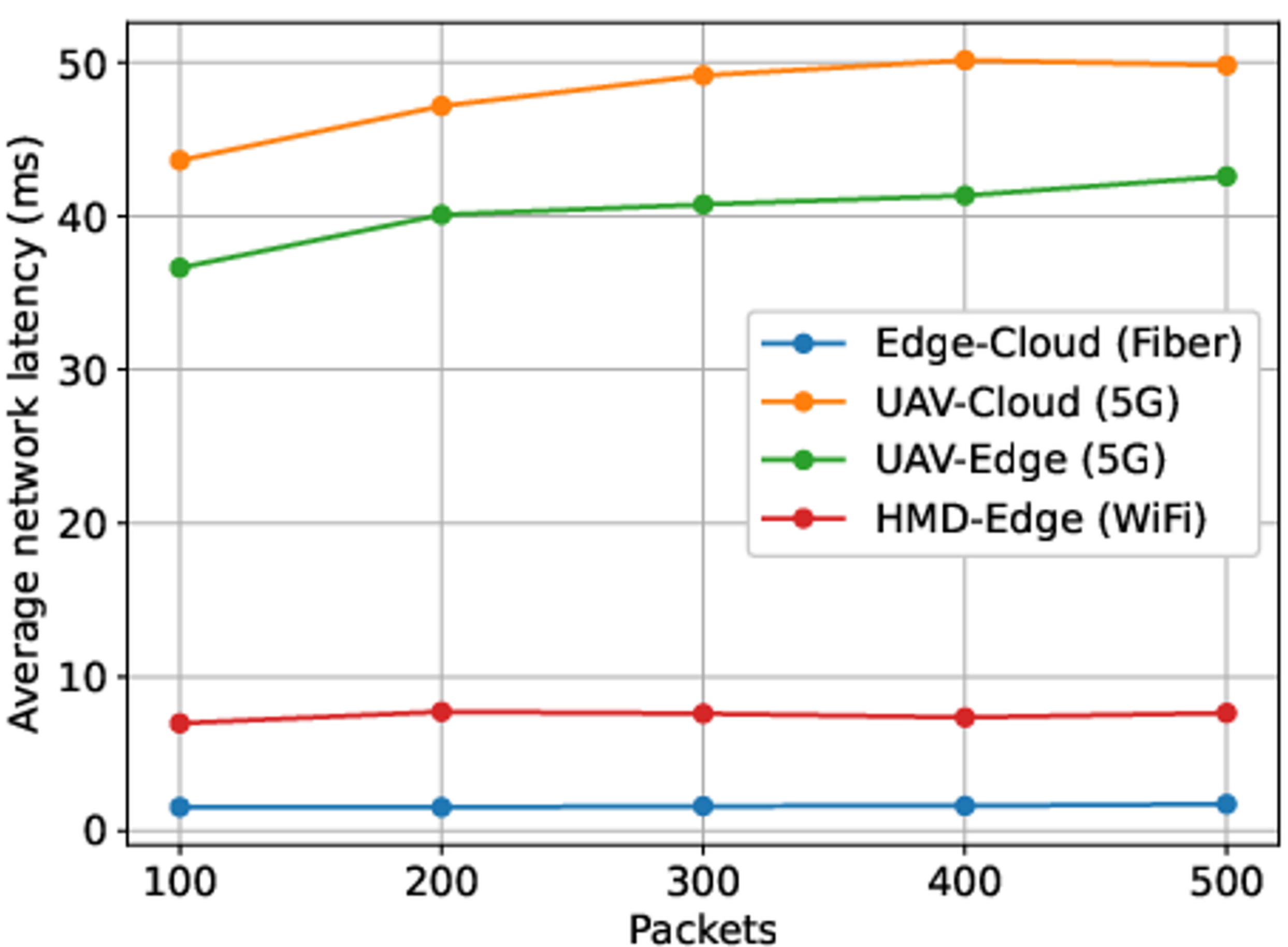}
    \caption{\textcolor{blue}{Network latency between components in the architecture}}
    \label{delays}
    \vspace{-0.5cm}
\end{figure}

\begin{table}[H]
\caption{Testbed's parameters and values.}
\label{table:1}
\begin{tabular}{p{0.5\linewidth}p{0.4\linewidth}}
\toprule
Parameter & Value \\
\midrule
Wifi (upload/download) & 100Mbps/200Mbps \\
5G (upload/download) & 50Mbps/200Mbps \\
Ethernet connection & 900Mbps/800Mbps \\
Edge Server CPU & 8 cores @ 2.5GHz \\
Edge Server memory & 16GB \\
Onboard computer memory & 8GB \\
Onboard CPU & 4 cores @ 1.5GHz \\
Distance \ac{uav} to Server & 300m \\
Distance \ac{vr} \ac{hmd} to Edge Server & 100m \\
\ac{vr} headset & Oculus Quest 2 \\
\bottomrule
\end{tabular}
\end{table}
The network latencies of the testbed are illustrated in Fig.~\ref{delays}. This figure provides a visual representation of the network latency and connection types among communicating devices within our testbed, encompassing edge to cloud, \ac{uav} to cloud, \ac{uav} to edge server, and \ac{hmd} to edge connections. These latency values delineate the spatial distribution of the devices relative to each other and their respective network connection types.

The highest latency, averaging 48ms, is observed between the \ac{uav} and the cloud server. This primarily stems from the \ac{uav}'s mobility and the resultant disruption of the 5G communication link due to frequent handovers at high altitudes. Conversely, latency is slightly lower between the \ac{uav} and the edge server, owing to the closer proximity of the edge server to the \ac{uav}. Notably, significantly lower delays are observed between the \ac{hmd} and the edge server, attributed to the stationary nature of the \ac{hmd} compared to the \ac{uav}, as well as its proximity to the edge server. The lowest latency, averaging 2ms, is noted between the edge server and the cloud server, which can be attributed to their direct fiber connection.

\subsection{Prompts and output}

At first, the first \ac{llm} is prompted by annotations and objects from the \ac{uav} and generates the following description for the video taken during our experiment: 

\texttt{The image depicts a large red building with a flat roof, surrounded by snow-covered trees and a snow-covered ground. There are two people in the foreground, one of them is holding a camera, and the other appears to be flying a drone.}

Thereafter, the second \ac{llm} is tasked with generating code to render the description in a \ac{3d} manner, based on the previous description provided by the first \ac{llm}, as shown in the following prompt. Notably, the prompt emphasizes the exclusion of external models such as \ac{gltf} and \ac{glb}:

\texttt{Generate A-Frame elements starting with 'a-' to accomplish the following instruction while meeting the conditions below.}

\textbf{Conditions:}

\begin{quote}
\texttt{- Do not use \texttt{a-assets} or \texttt{a-light}.\\
- Avoid using scripts.\\
- Do not use \ac{gltf}, \ac{glb} models.\\
- Do not use external model links.\\
- Provide animation.\\
- Use high-quality detailed models.\\
- If animation setting is requested, use the animation component instead of the \texttt{<a-animation>} element.\\
- If the background setting is requested, use the \texttt{<a-sky>} element instead of the background component.\\
- Provide the result in one code block.}
\end{quote}

 \textbf{Instruction:} \begin{quote}
\texttt{You are an assistant that teaches me Primitive Element tags for A-Frame version 1.4.0 and later. Create a 'Description from first \ac{llm}'.}
\end{quote}

\begin{figure}[H]
    \centering
    \includegraphics[width=0.5\textwidth]{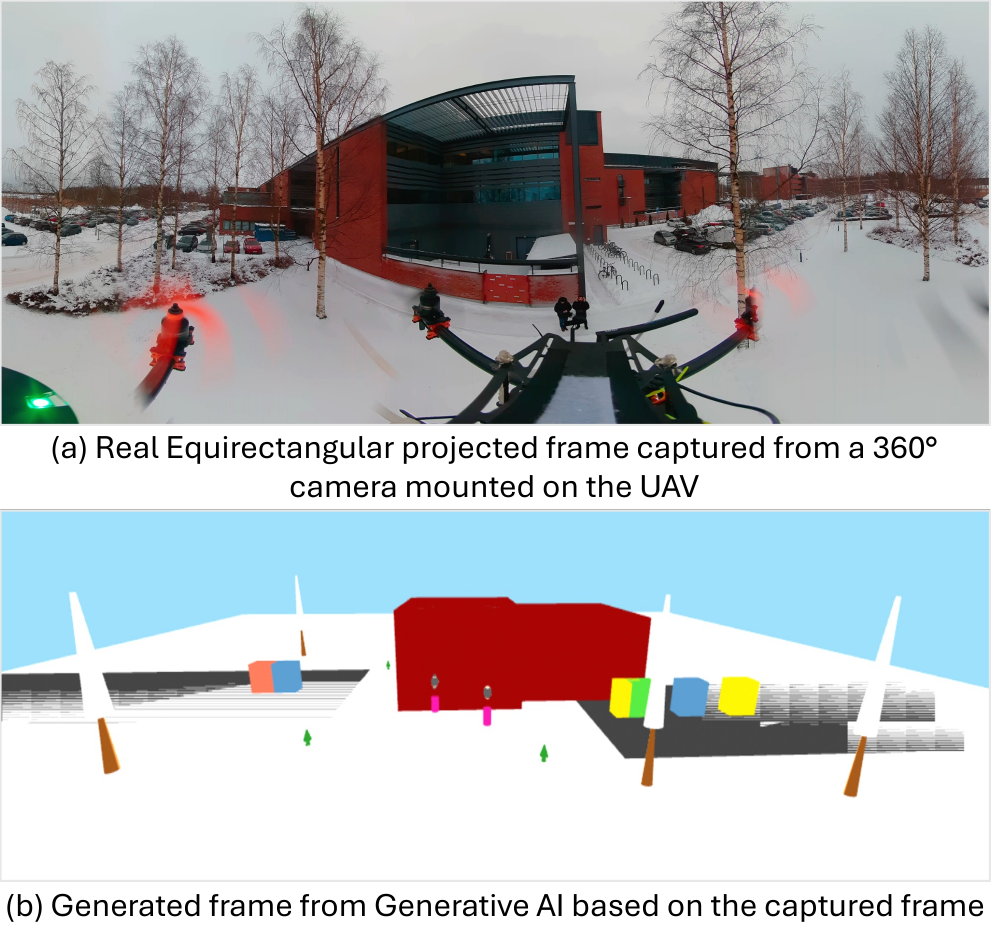}
    \caption{\textcolor{blue}{Generated \ac{3d} view against real captured image}}
    \label{fig:generated image}
\end{figure}

As an output, the rendered code is represented in Fig.~\ref{fig:generated image}, which shows both  the captured 360$^{\circ}$ frames from the camera 
\textcolor{blue}{in Fig.~\ref{fig:generated image} (a)} and \ac{3d} content generated based on \ac{html} code created by the \ac{llm} \textcolor{blue}{in Fig .~\ref{fig:generated image} (b)}.
The view undergoes transformation onto a spherical projection to align with the user's \ac{fov}.

\begin{figure*}[ht]
    \centering
    \begin{subfigure}[b]{0.554\textwidth}
        \includegraphics[width=\textwidth]{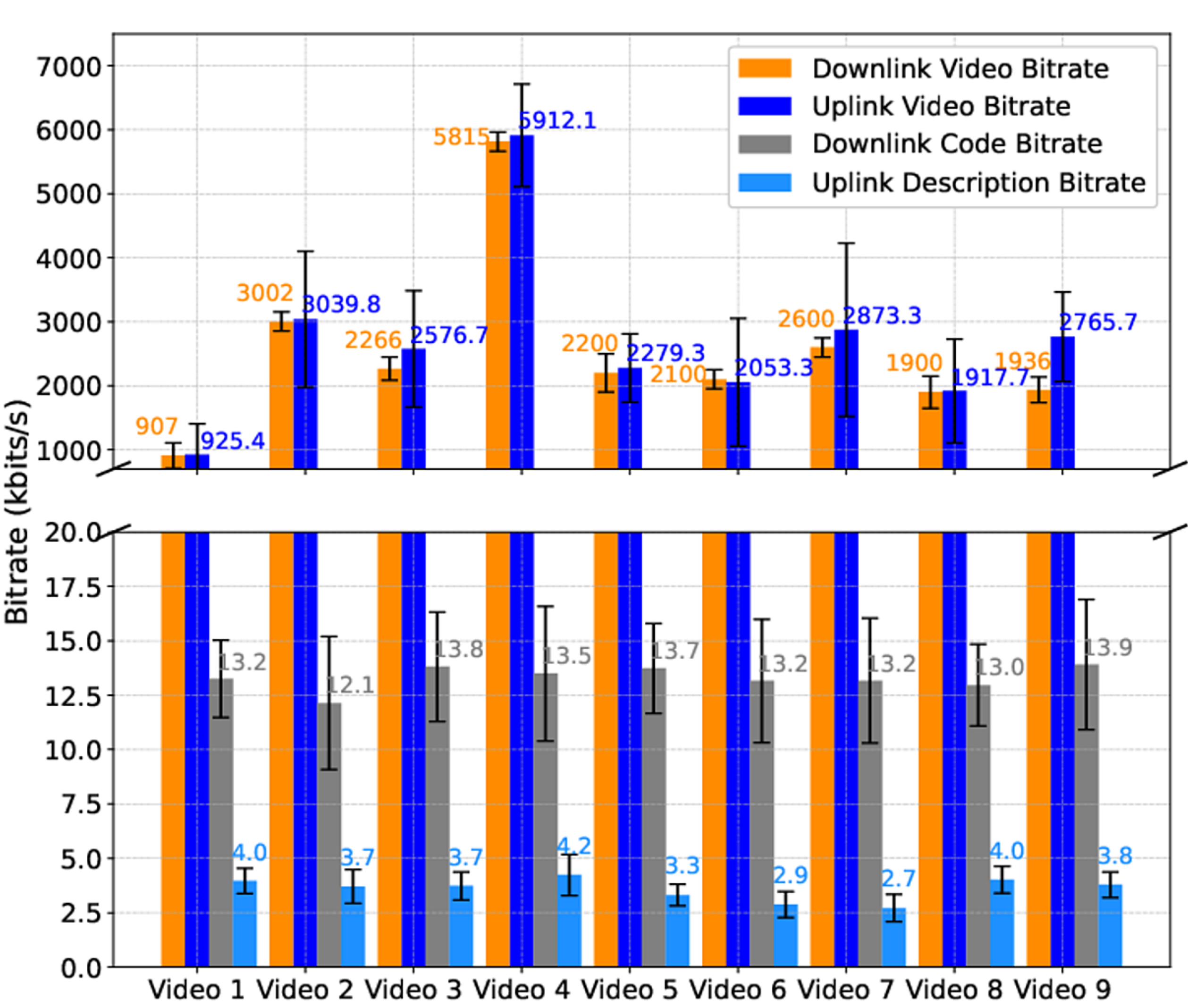}
        \caption{Average Download/Upload Bitrate for Video and generated Description/Code with Standard Deviation}
        \label{bandwidth}
    \end{subfigure}
    \hfill 
    \begin{subfigure}[b]{0.44\textwidth}
        \includegraphics[width=\textwidth]{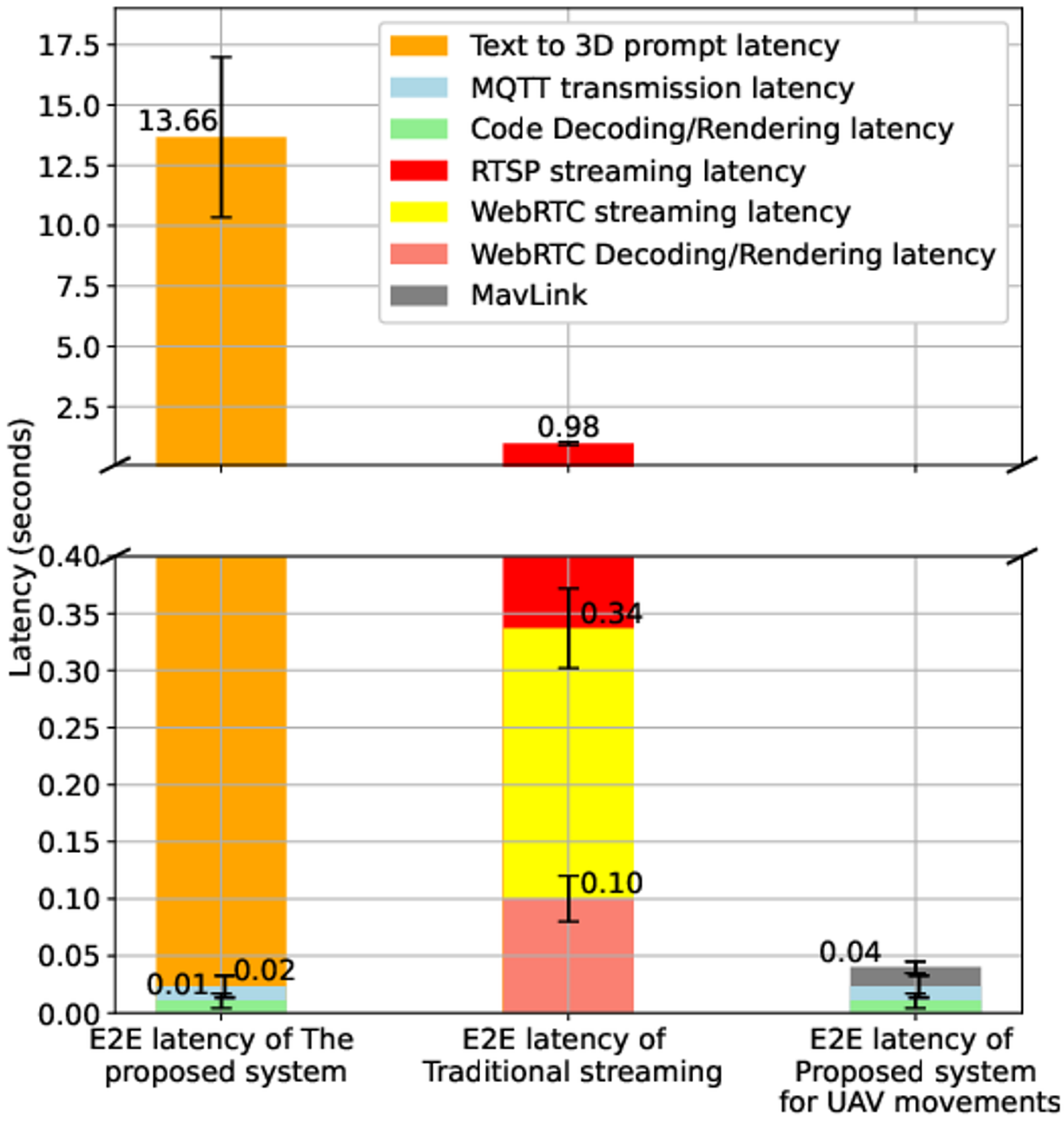}
        \caption{Average latency of both traditional and proposed systems for 360$^{\circ}$ video streaming}
        \label{delays-system}
    \end{subfigure}
    \caption{\textcolor{blue}{Comparison of bandwidth requirements and latency scores.}}
    \label{combined-figures}
\end{figure*}

\subsection{Experimental measurements}
To measure bandwidth consumption during the upload phase of traditional video streaming, we recorded the bitrate using the FFmpeg~\textcolor{blue}{\footnote{\href{https://www.ffmpeg.org/}{https://www.ffmpeg.org/}}} command at the \ac{uav}. Simultaneously, we utilized the \ac{webrtc} statistics \ac{api} at the \ac{hmd} level. For our proposed method, we calculated the average size of the description sent from the \ac{uav} and the size of the received \ac{llm}-generated code at the \ac{hmd}. In both traditional and our proposed systems, we analyzed various delays, including the \ac{e2e} traditional video streaming latency (L). This latency is constituted by the \ac{rtsp} video stream from the \ac{uav} to the edge server, the \ac{webrtc} stream from the edge server to the \ac{hmd}, and the frame rendering delays at the \ac{hmd}, as depicted in Equation~\eqref{eq:trad}.
\begin{equation}
\text{E2EL}_{\text{Traditional}} = \text{L}_{\text{RTSP}} + \text{L}_{\text{WebRTC}} + \text{L}_{\text{Rendering}}
\label{eq:trad}
\end{equation}
The constituting latencies in this latter case have been measured at the edge server, namely for the \ac{rtsp} streaming, and at the \ac{hmd} for \ac{webrtc} streaming and rendering. \textcolor{blue}{Our method mainly} encompasses the latency of text prompt to \ac{3d} WebXR code generation from the two \acp{llm} used, the code transmission through \ac{mqtt}, and WebXR code rendering. Considering that we can achieve real-time 30 \ac{fps} object detection 
using the onboard computer and that captioning is only applied to one frame out of 30, we consider the object detection latency negligible. The \ac{e2e} latency (\(\text{L}_{\text{Our Method}}\)) can be expressed as shown in Equation~\eqref{eq:our_method_latency}.
\begin{equation}
\text{E2EL}_{\text{Our Method}} = \text{L}_{\text{Text to Code}} + \text{L}_{\text{MQTT}} + \text{L}_{\text{Code Rendering}}
\label{eq:our_method_latency}
\end{equation}

The constituent latencies were measured by capturing timestamps from the sending device to the moment the response is generated and dispatched back to the sender, thus representing the round-trip latency. To approximate the one-way latency, this round-trip latency was halved. Additionally, we gauged the latency involved in transmitting \ac{uav} positions and synchronizing camera movement by leveraging the TIMESYNC protocol. It is noteworthy that all measurements presented herein reflect the average latency across the transmitted packet count.

\subsection{Results and analysis}
To analyze our system, we measured both upload and download bandwidth consumption, as well as the latency required to stream equirectangular frames of the test videos under consideration. Subsequently, we compared these metrics with those associated with traditional video streaming, focusing on our method, which involves generating virtual \ac{3d} objects based on \ac{llm} through semantic annotations, as illustrated in Fig.~\ref{combined-figures}.

In the case of traditional video streaming, the measured uplink and downlink bandwidth shown in Fig.~\ref{bandwidth} represent the average size of data streamed from the \ac{uav} to the user. Using our streaming method involves the uplink handling of semantic annotations and captioning descriptions sent by the drone. From the downlink perspective, it represents the size of the generated code to produce a \ac{3d} virtual animation mimicking the real environment for the \ac{vr} user.

We observe that in traditional video streaming, the uplink and downlink are almost similar, with the downlink being slightly lower due to \ac{webrtc} adapting to the available network bandwidth and latency. This difference in bandwidth requirements is attributed to the complexity of frames based on \textcolor{blue}{the content of each video}.  A similar variation is present in the uplink annotation streaming from our method, which is also due to the different descriptions and detected objects from the videos' frames. The downlink, on the other hand, consistently has the same size, attributed to the code and output of the \ac{llm}, which is restricted by the prompt to a predetermined size of generated tokens.

In summary, the bandwidth analysis reveals that our proposed method requires only a few kilobits per second (kbps), with a maximum of 13.9kbps in the uplink for the 9 videos. In comparison, traditional video streaming demands 5.9Mbps, while in the download, our method needs 4kbps, contrasting with 5.8Mbps in traditional streaming resulting in a reduction of approximately 99.93\%. 

As observed in the latency analysis depicted in Fig~\ref{delays-system}, our method exhibits a latency approximately 13 times higher, with an average latency of 13.66 seconds, compared to 980ms in traditional streaming. This increase is primarily attributed to the prompt-to-token latency of the large-sized \acp{llm}, as well as network latencies, given that the \acp{llm} are situated in the cloud. Additionally, we have suggested an approach to reduce these delays by creating a smaller version of the \acp{llm}. It is worth highlighting that the decoding and rendering code for animated \ac{3d} objects takes significantly less time than processing captured images, with an average duration of 10ms compared to 100ms per 30 frames. Since we continuously update the virtual camera view position according to the streamed \ac{uav} positions with a delay of 40ms, the \ac{uav} control is not affected, as static objects will already be presented.

\begin{figure*}[h]

\begin{tikzpicture}
\hspace{-0.2cm} 

    \begin{axis}[
        ybar,
        bar width=0.4cm,  
        width=19cm,  
        height=8cm,  
        enlarge x limits=0.09,  
        legend style={
            at={(0.5, 1.05)},  
            anchor=south,  
            legend columns=3,  
            font=\small  
        },
        ylabel={\textcolor{blue}{\ac{bert} semantic similarity score}},
        xlabel={\textcolor{blue}{Video Number}},
        ylabel style={yshift=-1.2em, font=\small},  
        xlabel style={yshift=-0.8ex, font=\small},  
        symbolic x coords={1,2,3,4,5,6,7,8,9, 10},  
        xtick={1,2,3,4,5,6,7,8,9, 10},  
        xticklabels={1,2,3,4,5,6,7,8,9, 10},  
        ymin=0,
        ymax=1,
        nodes near coords,  
        every node near coord/.append style={
            font=\small,  
            yshift=-0.15em,  
            anchor=south  
        }
    ]
        \addplot[
            ybar,
            fill=blue,  
            fill opacity=0.3,
            nodes near coords,
            every node near coord/.append style={
                font=\small,
                xshift=-0.3em, 
                yshift=-0.02em, 
                anchor=south,
                text=blue,
                text opacity=1,
            }
        ]
             coordinates {(1,0.8254102658369376) (2,0.787088750460761) (3,0.623419953174213) (4, 0.601679377436844) (5,0.694056113815522) (6,0.7772322810100852) (7,0.6694825250830971) (8,0.7669069653476257) (9,0.8329111145244158) (10,0.8858416251319354)};
        
        \addplot[
            ybar,
            fill=red,  
            fill opacity=0.25,
            nodes near coords,
            every node near coord/.append style={
                font=\small,
                anchor=south,
                text=red,
                text opacity=1,
            }
        ] coordinates {(1,0.7796725715050203) (2,0.6926982289895598) (3,0.555295166437075) (4, 0.7180371881058016) (5,0.6209982430809817) (6,0.7318030166280459) (7,0.6886408379336151) (8,0.703684418002533) (9,0.8245002371561927) (10,0.8047367878623054)};
        
        \addplot coordinates {(1,0.8716732662650761) (2,0.8580468459379859) (3,0.7633648863656888) (4, 0.7898976196131304) (5,0.7333237143026153) (6,0.8021394896909184) (7,0.6944647045765186) (8,0.7934809796123743) (9,0.9483814071332302) (10,0.9063121111987439)};
        
        \legend{Our captioning, \ac{vit} over GPT-2 captioning, GPT-4o captioning}
    \end{axis}
\end{tikzpicture}
  \caption{\textcolor{blue}{Description similarity between generated captions compared to human captions}}
  \label{fig:caption}
  \vspace{-0.3cm}
\end{figure*}

\begin{figure}[h]
  \centering
  \begin{tikzpicture}
    \begin{axis}[
      ybar,
      bar width=0.5cm,
      xlabel={Video Number},
      ylabel={\textcolor{blue}{\ac{bert} semantic similarity score}},
      symbolic x coords={1, 2, 3, 4, 5, 6, 7, 8, 9, 10},
      xtick=data,
      nodes near coords,
      nodes near coords align={vertical},
      ymin=0,
      ymax=1,
      ]
      \addplot[fill=blue!30] coordinates {
        (1, 0.7814)
        (2, 0.6770)
        (3, 0.55)
        (4, 0.6090)
        (5, 0.444)
        (6, 0.4286)
        (7, 0.7467)
        (8, 0.8329)
        (9, 0.7329)
        (10, 0.7570)
      };
    \end{axis}
  \end{tikzpicture}
  \caption{\textcolor{blue}{Description similarity between generated frames compared to real frames}}
  \label{fig:video_similarity_bar_chart}
  \vspace{-0.3cm}
\end{figure}
\subsection{Validation}
To validate our proposed system, \textcolor{blue}{we evaluated the semantic similarity between the output of the first \ac{llm} used in our framework and human-annotated descriptions of the 10 video frames. We employed \ac{bert}, a widely recognized method for assessing the degree of semantic textual similarity. Additionally, we compared our results with those obtained from captioning methods based on transformer architectures, namely \ac{vit} over GPT-2, and GPT-4o. Fig.~\ref{fig:caption} illustrates the similarity scores between human-generated captions and those produced by our method, \ac{vit} over \ac{gpt}-2, and \ac{gpt}-4o for a randomly selected frame from each video. The results demonstrate the effectiveness of our method, achieving an average similarity of 71\% with human captions. Notably, our approach particularly excels with the 10th video, which incorporates \ac{uav} sensorial data. While the \ac{mllm} GPT-4o might offer the optimal solution due to its training on a broader range of data, it necessitates streaming entire frames, incurring substantially higher bandwidth consumption compared to our approach.}

\textcolor{blue}{Subsequently}, we compared the descriptions generated by a \ac{vlm}, specifically GPT-4, for both the virtual \ac{3d} frames (generated by the second \ac{llm}) and the equirectangular frames of the ten videos. As direct frame comparison using traditional metrics like \ac{psnr} is not suitable, we employed \ac{bert}-based average semantic comparison, with the results depicted in Fig.~\ref{fig:video_similarity_bar_chart}. Notably, the maximum \ac{bert} similarity score signifies the highest probability of 1.

Overall, the results indicate a satisfactory representation of code-based descriptions. A maximum matching score of 83\% was observed for the 8th video (Buddhist temple), while the minimum score of 43\% was associated with the 6th video (park with animals and a lake). The lower representation quality in videos 3, 5, and 6 can be attributed to their inherent complexity and the presence of numerous objects. The A-Frame WebXR framework used in this study generates \ac{3d} representations using basic geometric shapes, potentially limiting its ability to accurately recreate such intricate scenes. Moreover, the utilized \acp{llm} were not fine-tuned for expertise in the WebXR framework. We also explored Code LLaMA\footnote{\url{https://ai.meta.com/blog/code-llama-large-language-model-coding/}}, designed specifically for coding tasks; however, its generated code was notably weaker compared to that of \ac{gpt}-4.

\section{Challenges \& Open Research Directions}

\subsection{Multi-user \& Scalability} 
Scalability for such applications as the one proposed in the use case is quite challenging since the response from an \ac{llm} when prompted by multiple tasks might be degraded by up to 3\% less accuracy for 50 simultaneous prompts~\cite{maatouk2023teleqna}. A scalable 6G network is needed in order to accommodate a large number of immersive users, while dynamically being able to \textcolor{blue}{adjust} its resources and services to serve varying demands without compromising performance, reliability, or user experience.

\subsection{Latency \& Real-time Processing}
In order to realize a fully immersive experience through \ac{ios}, the utilized \acp{llm} should be capable of processing and interpreting vast amounts of sensory data in real-time, facilitating seamless human-machine interactions. Additionally, they need to be optimized for edge computing architectures to ensure that data processing is as close to the source as possible. The challenge in achieving real-time processing in 6G lies in minimizing latency to the extent that the delay is imperceptible to humans or sensitive systems, which requires major advancements in network infrastructure, edge computing capabilities, and \acp{llm}.


\subsection{Edge Computation Limitations}
Deploying \acp{llm} on \acp{ue} or small edge servers presents challenges due to the computational demands of these models. \acp{llm} require substantial processing power and memory resources. However, mobile devices often have limited resources compared to desktop computers or servers. Consequently, running \acp{llm} on \acp{ue} may lead to slower inference times and reduced overall performance. Additionally, their typical constraint to fewer than 7 billion parameters frequently results in decreased response quality, with distortion being a common occurrence in tasks such as image generation \cite{zhong2023mobile}.

\subsection{Energy consumption}
\acp{llm} are computationally intensive and can consume a significant amount of power during inference. Given the limited battery capacity of mobile devices, running \acp{llm} for extended periods can quickly drain the battery. This limitation significantly impacts the practicality and usability of \acp{llm} on mobile devices, especially when offline or in situations without immediate access to power sources.

\subsection{Integration \& Interoperability}
The seamless interoperability of \ac{ios} and \acp{llm} among a vast array of devices, technologies, and protocols constitutes a main challenge for future 6G networks. This integration will require a sophisticated orchestration of network components to ensure that the high-speed, low-latency, accuracy, and reliability are not compromised. This necessitates the development of adaptive network architectures that are capable of handling the diverse demands of sensory-data processing and \ac{ai} interactions within a large number of users. 


\section{Conclusion} \label{conclusion}
\textcolor{blue}{This paper has established a foundational framework for integrating \acp{llm} with the \ac{ios} within the context of 6G networks.} We have defined the key principles of \ac{ios} and presented promising use cases that showcase the potential of \acp{llm} in enabling low-latency, multi-sensory communication experiences. Within these use cases, we have explored the application of \acp{llm} as effective compressors and showcased a practical implementation on a real testbed, leveraging generative \ac{ai} for the \ac{ios}. The measurement methodologies and analysis of the proposed system have been meticulously detailed and benchmarked against traditional approaches to multi-sensory data transmission. Our results demonstrate that \acp{llm} can achieve significant bandwidth savings; however, their response latency currently presents a challenge for real-time applications. To alleviate this limitation, we have designed and presented an approach focused on fine-tuning \acp{llm} and deploying them closer to the user. Looking ahead, we intend to investigate the use of fine-tuned \acp{llm} directly on \acp{uav} as an alternative to conventional captioning and object detection methods, potentially enhancing the sensory experience within \ac{ios} applications.



\bibliographystyle{IEEEtran}
\bibliography{IEEEabrv,ref}

\begin{thebibliography}{10}
\providecommand{\url}[1]{#1}
\csname url@samestyle\endcsname
\providecommand{\newblock}{\relax}
\providecommand{\bibinfo}[2]{#2}
\providecommand{\BIBentrySTDinterwordspacing}{\spaceskip=0pt\relax}
\providecommand{\BIBentryALTinterwordstretchfactor}{4}
\providecommand{\BIBentryALTinterwordspacing}{\spaceskip=\fontdimen2\font plus
\BIBentryALTinterwordstretchfactor\fontdimen3\font minus \fontdimen4\font\relax}
\providecommand{\BIBforeignlanguage}[2]{{%
\expandafter\ifx\csname l@#1\endcsname\relax
\typeout{** WARNING: IEEEtran.bst: No hyphenation pattern has been}%
\typeout{** loaded for the language `#1'. Using the pattern for}%
\typeout{** the default language instead.}%
\else
\language=\csname l@#1\endcsname
\fi
#2}}
\providecommand{\BIBdecl}{\relax}
\BIBdecl

\bibitem{9143472}
M.~Melo \emph{et~al.}, ``Do multisensory stimuli benefit the virtual reality experience? a systematic review,'' \emph{IEEE Transactions on Visualization and Computer Graphics}, vol.~28, no.~2, pp. 1428--1442, 2022.

\bibitem{pyun2022materials}
K.~R. Pyun \emph{et~al.}, ``Materials and devices for immersive virtual reality,'' \emph{Nature Reviews Materials}, vol.~7, no.~11, pp. 841--843, 2022.

\bibitem{fettweis2014tactile}
G.~Fettweis \emph{et~al.}, ``The tactile internet-itu-t technology watch report,'' \emph{Int. Telecom. Union (ITU), Geneva}, 2014.

\bibitem{akyildiz2023mulsemedia}
I.~F. Akyildiz \emph{et~al.}, ``Mulsemedia communication research challenges for metaverse in {6G} wireless systems,'' \emph{arXiv preprint arXiv:2306.16359}, 2023.

\bibitem{deletang2023language}
G.~Del{\'e}tang \emph{et~al.}, ``Language modeling is compression,'' \emph{arXiv preprint arXiv:2309.10668}, 2023.

\bibitem{NEURIPS2020_1457c0d6}
\BIBentryALTinterwordspacing
T.~Brown \emph{et~al.}, ``Language models are few-shot learners,'' in \emph{Advances in Neural Information Processing Systems}, H.~Larochelle, M.~Ranzato, R.~Hadsell, M.~Balcan, and H.~Lin, Eds., vol.~33.\hskip 1em plus 0.5em minus 0.4em\relax Curran Associates, Inc., 2020, pp. 1877--1901. [Online]. Available: \url{https://proceedings.neurips.cc/paper_files/paper/2020/file/1457c0d6bfcb4967418bfb8ac142f64a-Paper.pdf}
\BIBentrySTDinterwordspacing

\bibitem{zhang2024mm}
D.~Zhang, Y.~Yu, C.~Li, J.~Dong, D.~Su, C.~Chu, and D.~Yu, ``Mm-llms: Recent advances in multimodal large language models,'' \emph{arXiv e-prints}, pp. arXiv--2401, 2024.

\bibitem{ranasinghe2012tongue}
N.~Ranasinghe \emph{et~al.}, ``Tongue mounted interface for digitally actuating the sense of taste,'' in \emph{2012 16th international symposium on wearable computers}.\hskip 1em plus 0.5em minus 0.4em\relax IEEE, 2012, pp. 80--87.

\bibitem{panagiotakopoulos2022digital}
D.~Panagiotakopoulos \emph{et~al.}, ``Digital scent technology: Toward the internet of senses and the metaverse,'' \emph{IT Professional}, vol.~24, no.~3, pp. 52--59, 2022.

\bibitem{Choi_2019}
Y.~E. Choi, ``A survey of 3d audio reproduction techniques for interactive virtual reality applications,'' \emph{IEEE Access}, vol.~7, pp. 26\,298--26\,316, 2019.

\bibitem{xing2024survey}
J.~Xing \emph{et~al.}, ``A survey of efficient fine-tuning methods for vision-language models—prompt and adapter,'' \emph{Computers \& Graphics}, 2024.

\bibitem{yuan2023tinygptv}
Z.~Yuan \emph{et~al.}, ``Tinygpt-v: Efficient multimodal large language model via small backbones,'' 2023.

\bibitem{lo2018edge}
W.-C. Lo, C.-Y. Huang, and C.-H. Hsu, ``Edge-assisted rendering of 360 videos streamed to head-mounted virtual reality,'' in \emph{2018 IEEE International Symposium on Multimedia (ISM)}.\hskip 1em plus 0.5em minus 0.4em\relax IEEE, 2018, pp. 44--51.

\bibitem{taleb2022vr}
T.~Taleb \emph{et~al.}, ``Vr-based immersive service management in b5g mobile systems: A uav command and control use case,'' \emph{IEEE Internet of Things Journal}, vol.~10, no.~6, pp. 5349--5363, 2022.

\bibitem{chen2021optimized}
X.~Chen, D.~Wu, and I.~Ahmad, ``Optimized viewport-adaptive 360-degree video streaming,'' \emph{CAAI Transactions on Intelligence Technology}, vol.~6, no.~3, pp. 347--359, 2021.

\bibitem{yi2020analysis}
J.~Yi, M.~R. Islam, S.~Aggarwal, D.~Koutsonikolas, Y.~C. Hu, and Z.~Yan, ``An analysis of delay in live 360° video streaming systems,'' in \emph{Proceedings of the 28th ACM International Conference on Multimedia}, 2020, pp. 982--990.

\bibitem{park2023omnilive}
S.~Park, Y.~Cho, H.~Jun, J.~Lee, and H.~Cha, ``Omnilive: Super-resolution enhanced 360 video live streaming for mobile devices,'' in \emph{Proceedings of the 21st Annual International Conference on Mobile Systems, Applications and Services}, 2023, pp. 261--274.

\bibitem{gao2024low}
N.~Gao, J.~Zhou, G.~Wan, X.~Hua, T.~Bi, and T.~Jiang, ``Low-latency vr video processing-transmitting system based on edge computing,'' \emph{IEEE Transactions on Broadcasting}, 2024.

\bibitem{de2024scalable}
M.~De~Fr{\'e}, J.~van~der Hooft, T.~Wauters, and F.~De~Turck, ``Scalable mdc-based volumetric video delivery for real-time one-to-many webrtc conferencing,'' in \emph{Proceedings of the 15th ACM Multimedia Systems Conference}, 2024, pp. 121--131.

\bibitem{uson2024untethered}
J.~Us{\'o}n, C.~Cort{\'e}s, V.~Mu{\~n}oz, T.~Hernando, D.~Berj{\'o}n, F.~Mor{\'a}n, J.~Cabrera, and N.~Garc{\'\i}a, ``Untethered real-time immersive free viewpoint video,'' in \emph{Proceedings of the 16th International Workshop on Immersive Mixed and Virtual Environment Systems}, 2024, pp. 45--49.

\bibitem{xia2023wiservr}
L.~Xia, Y.~Sun, C.~Liang, D.~Feng, R.~Cheng, Y.~Yang, and M.~A. Imran, ``Wiservr: Semantic communication enabled wireless virtual reality delivery,'' \emph{IEEE Wireless Communications}, vol.~30, no.~2, pp. 32--39, 2023.

\bibitem{ahn2024dynamic}
S.~Ahn, H.-J. Yim, Y.~Lee, and S.-I. Park, ``Dynamic and super-personalized media ecosystem driven by generative ai: Unpredictable plays never repeating the same,'' \emph{IEEE Transactions on Broadcasting}, 2024.

\bibitem{chen2024cross}
M.~Chen, M.~Liu, C.~Wang, X.~Song, Z.~Zhang, Y.~Xie, and L.~Wang, ``Cross-modal graph semantic communication assisted by generative ai in the metaverse for 6g,'' \emph{Research}, vol.~7, p. 0342, 2024.

\bibitem{du2023yolo}
B.~Du, H.~Du, H.~Liu, D.~Niyato, P.~Xin, J.~Yu, M.~Qi, and Y.~Tang, ``Yolo-based semantic communication with generative ai-aided resource allocation for digital twins construction,'' \emph{IEEE Internet of Things Journal}, 2023.

\bibitem{sharma2023uav}
M.~K. Sharma, C.-F. Liu, I.~Farhat, N.~Sehad, W.~Hamidouche, and M.~Debbah, ``Uav immersive video streaming: A comprehensive survey, benchmarking, and open challenges,'' \emph{arXiv preprint arXiv:2311.00082}, 2023.

\bibitem{meng2023dtuav}
W.~Meng, Y.~Yang, J.~Zang, H.~Li, and R.~Lu, ``Dtuav: a novel cloud--based digital twin system for unmanned aerial vehicles,'' \emph{Simulation}, vol.~99, no.~1, pp. 69--87, 2023.

\bibitem{10437527}
N.~Sehad, X.~Tu, A.~Rajasekaran, H.~Hellaoui, R.~Jäntti, and M.~Debbah, ``Towards enabling reliable immersive teleoperation through digital twin: A uav command and control use case,'' in \emph{GLOBECOM 2023 - 2023 IEEE Global Communications Conference}, 2023, pp. 6420--6425.

\bibitem{degadwala2021image}
S.~Degadwala \emph{et~al.}, ``Image captioning using inception v3 transfer learning model,'' in \emph{2021 6th International Conference on Communication and Electronics Systems (ICCES)}.\hskip 1em plus 0.5em minus 0.4em\relax IEEE, 2021, pp. 1103--1108.

\bibitem{maatouk2023teleqna}
A.~Maatouk \emph{et~al.}, ``Teleqna: A benchmark dataset to assess large language models telecommunications knowledge,'' \emph{arXiv preprint arXiv:2310.15051}, 2023.

\bibitem{zhong2023mobile}
R.~Zhong \emph{et~al.}, ``Mobile edge generation: A new era to 6g,'' \emph{arXiv preprint arXiv:2401.08662}, 2023.

\end{thebibliography}

\begin{IEEEbiography}
[{\includegraphics[width=1in,height=2in,clip,keepaspectratio]{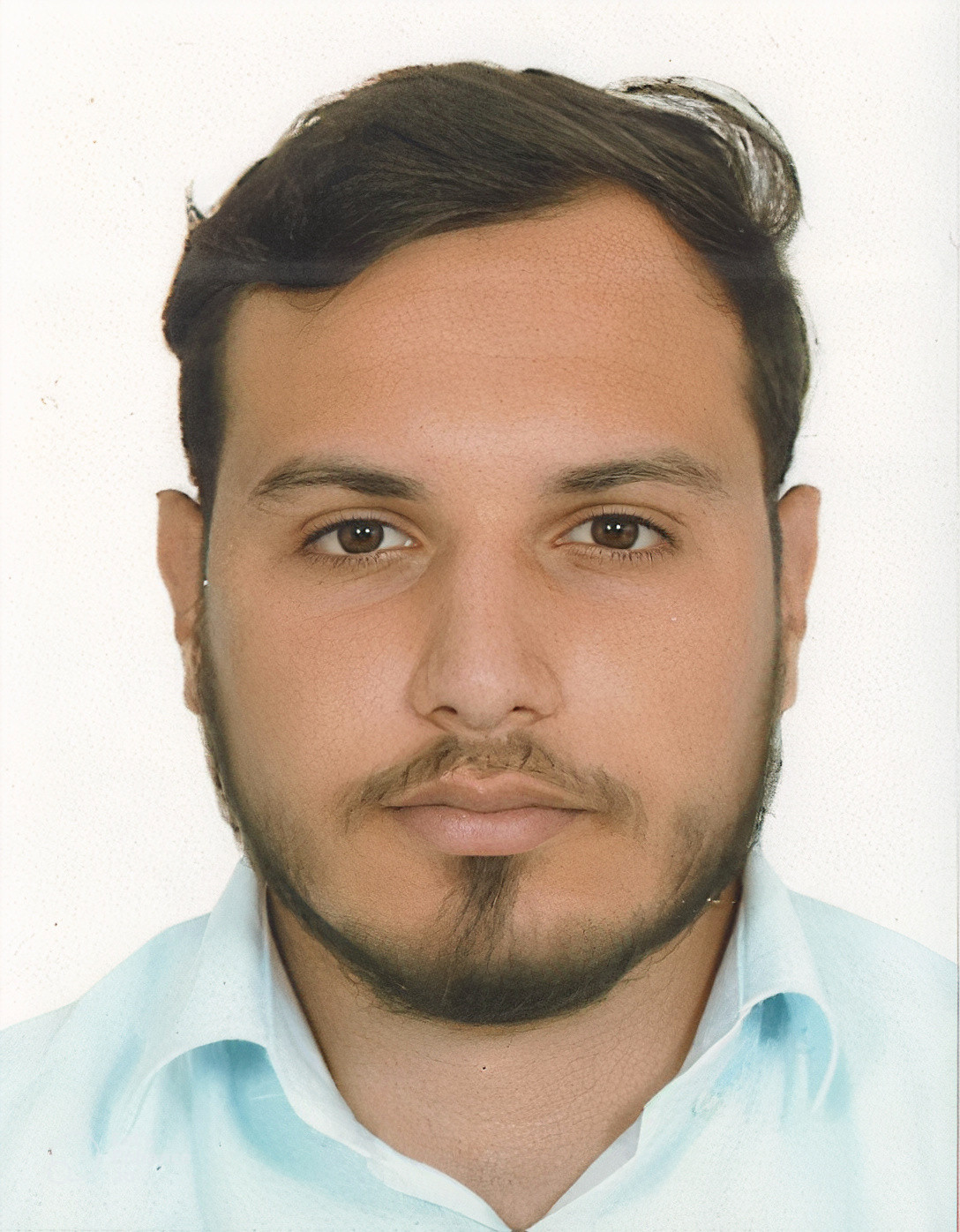}}]{Nassim Sehad} (nassim.sehad@aalto.fi) obtained  the Bachelor of Science (B.Sc)  diploma in the field of telecommunication in 2018 and the diploma of master in the field of networks, and telecommunication in September 2020, from the University of Sciences and Technology Houari Boumediene (U.S.T.H.B), Algiers, Algeria. Since 2020 to September 2021 he joined the MOSA!C laboratory at Aalto University Finland as an assistant researcher.
Since 2021 till now he joined the Department of Information and Communications Engineering (DICE), Aalto University, Finland, as a doctoral student.
His main research topics of interest are multi-sensory multimedia, IoT, cloud computing, networks and AI.
\end{IEEEbiography}
\begin{IEEEbiography}
[{\includegraphics[width=1in,height=2in,clip,keepaspectratio]{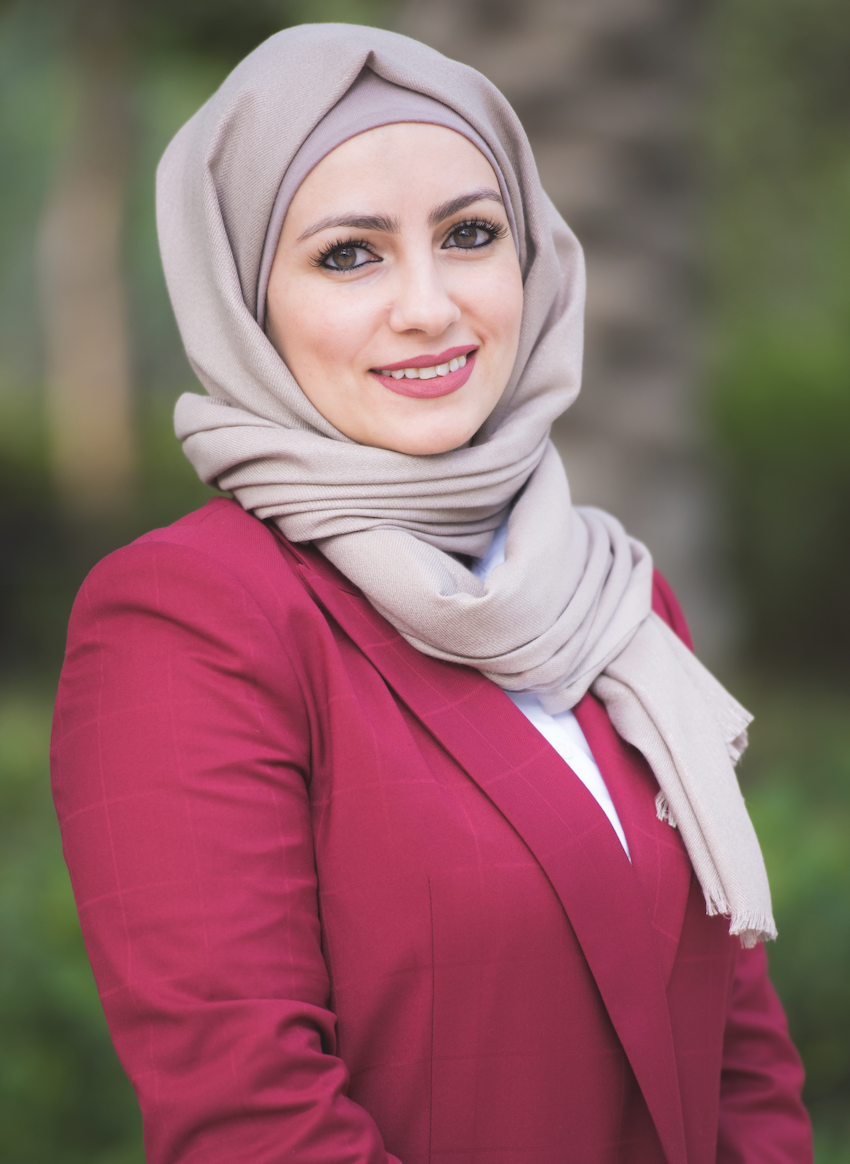}}]{Lina Bariah} (lina.bariah@ieee.org) Lina Bariah received the Ph.D. degree in communications engineering from Khalifa University, Abu Dhabi, UAE, in 2018. She is currently a Lead AI Scientist at Open Innovation AI, an Adjunct Professor at Khalifa University, and an Adjunct Research Professor, Western University, Canada. She was a Visiting Researcher with the Department of Systems and Computer Engineering, Carleton University, Ottawa, ON, Canada, in 2019, and an affiliate research fellow, James Watt School of Engineering, University of Glasgow, UK. She was a Senior Researcher at the technology Innovation institute. Dr. Bariah is a senior member of the IEEE. 
\end{IEEEbiography}
\vspace{-1cm}

\begin{IEEEbiography}[{\includegraphics[width=1in,height=1.25in,clip,keepaspectratio]{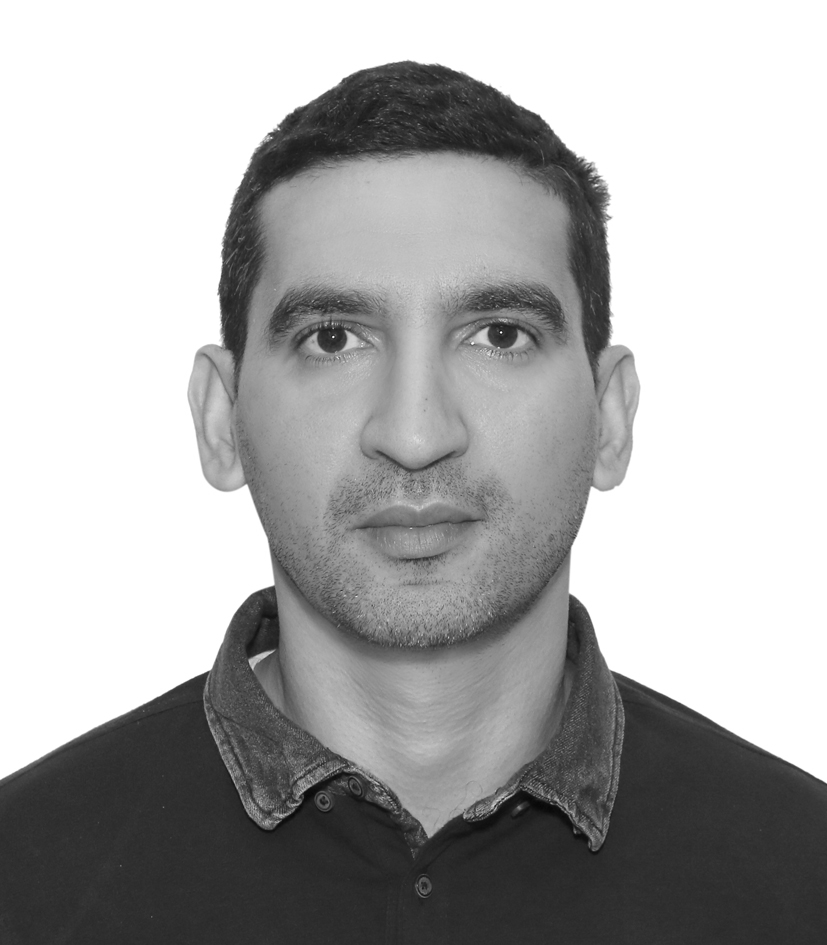}}]{Wassim Hamidouche} (wassim.hamidouche@tii.ae) is a Principal Researcher at Technology Innovation Institute (TII) in Abu Dhabi, UAE. He also holds the position of Associate Professor at INSA Rennes and is a member of the Institute of Electronics and Telecommunications of Rennes (IETR), UMR CNRS 6164. He earned his Ph.D. degree in signal and image processing from the University of Poitiers, France, in 2010. From 2011 to 2012, he worked as a Research Engineer at the Canon Research Centre in Rennes, France. Additionally, he served as a researcher at the IRT b$<>$com research Institute in Rennes from 2017 to 2022. He has over 180 papers published in the field of image processing and computer vision. His research interests encompass various areas, including video coding, the design of software and hardware circuits and systems for video coding standards, image quality assessment, and multimedia security.
\end{IEEEbiography}
\vspace{-1cm}

\begin{IEEEbiography}
[{\includegraphics[width=1in,height=1.25in,clip,keepaspectratio]{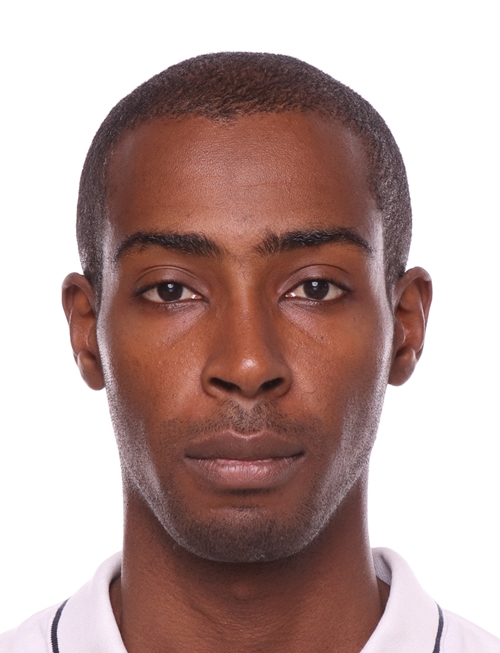}}]{Hamed Hellaoui} (hamed.hellaoui@nokia.com) received the Ph.D. degree in Computer Science from Ecole nationale Supérieure d'Informatique -Algeria- in 2021, and the Ph.D. degree in Communications and Networking from Aalto University -Finland- in 2022. He is currently a Senior Research Specialist at Nokia, Finland. He has been actively contributing to Nokia’s Home Programs on Research and Standardization related to 5GA/6G, as well as to several EU-funded projects. His research interests span diverse areas, including 5G and 6G communications, UAV, IoT, and machine learning.
\end{IEEEbiography}
\vspace{-1cm}

\begin{IEEEbiography}[{\includegraphics[width=1in,height=1.25in,clip,keepaspectratio]{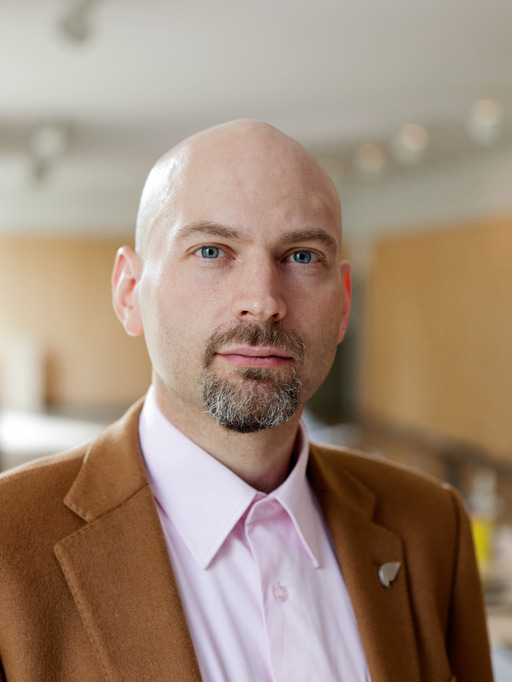}}]{Riku Jäntti} (Senior Member, IEEE) (M’02 - SM’07) (riku.jantti@aalto.fi) is a Full Professor of Communications Engineering at Aalto University School of Electrical Engineering, Finland. He received his M.Sc (with distinction) in Electrical Engineering in 1997 and D.Sc (with distinction) in Automation and Systems Technology in 2001, both from Helsinki University of Technology (TKK). Prior to joining Aalto in August 2006, he was professor pro tem at the Department of Computer Science, University of Vaasa. Prof. Jäntti is a senior member of IEEE. He has also been IEEE VTS Distinguished Lecturer (Class 2016). The research interests of Prof. Jäntti include machine type communications, disaggregated radio access networks, backscatter communications, quantum communications, and radio frequency inference.
\end{IEEEbiography}
\vspace{-1cm}
\begin{IEEEbiography}
[{\includegraphics[width=1in,height=2in,clip,keepaspectratio]{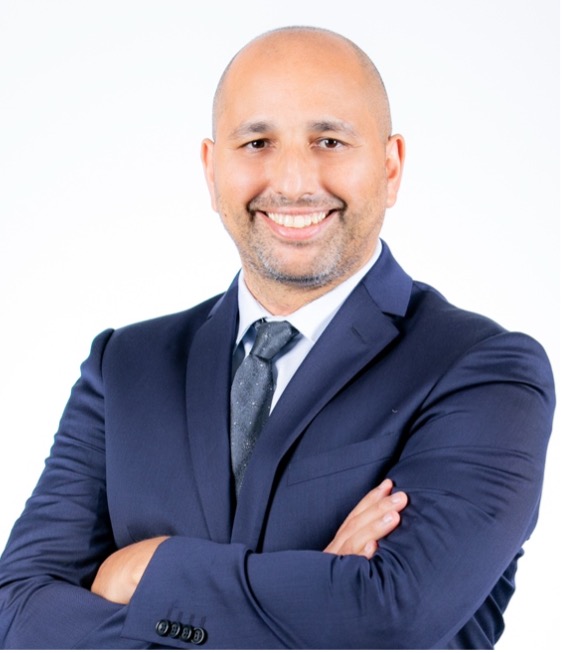}}]{Merouane Debbah} (merouane.debbah@ku.ac.ae) is a researcher, educator and technology entrepreneur. Over his career, he has founded several public and industrial research centers and start-ups, and is now Professor at Khalifa University of Science and Technology in Abu Dhabi and founding Director of the KU 6G research center. He is a frequent keynote speaker at international events in the field of telecommunication and AI. His research has been lying at the interface of fundamental mathematics, algorithms, statistics, information and communication sciences with a special focus on random matrix theory and learning algorithms. In the Communication field, he has been at the heart of the development of small cells (4G), Massive MIMO (5G) and Large Intelligent Surfaces (6G) technologies. In the AI field, he is known for his work on Large Language Models, distributed AI systems for networks and semantic communications. He received multiple prestigious distinctions, prizes and best paper awards (more than 35 best paper awards) for his contributions to both fields and according to research.com is ranked as the best scientist in France in the field of Electronics and Electrical Engineering. He is an IEEE Fellow, a WWRF Fellow, a Eurasip Fellow, an AAIA Fellow, an Institut Louis Bachelier Fellow and a Membre émérite SEE.
\end{IEEEbiography}





\end{document}